\documentclass{article} %
\usepackage{iclr2027_conference,times}
\iclrfinalcopy

\usepackage{amsmath,amsfonts,bm}

\def\eqref#1{equation~\ref{#1}}

\def\1{\bm{1}}

\DeclareMathAlphabet{\mathsfit}{\encodingdefault}{\sfdefault}{m}{sl}
\SetMathAlphabet{\mathsfit}{bold}{\encodingdefault}{\sfdefault}{bx}{n}

\usepackage{hyperref}
\usepackage{url}
\usepackage{graphicx}
\usepackage{float}
\usepackage{enumitem}
\usepackage{adjustbox}
\usepackage{booktabs,tabularx,colortbl}
\usepackage{array}
\usepackage{pifont}
\usepackage{makecell}
\usepackage{listings}
\usepackage{multirow}
\usepackage{microtype}
\usepackage{longtable,booktabs,array}
\usepackage{amsthm}
\theoremstyle{plain}
\newtheorem{proposition}{Proposition}
\newtheorem{corollary}{Corollary}
\theoremstyle{definition}
\newtheorem{definition}{Definition}
\newtheorem{assumption}{Assumption}
\theoremstyle{remark}
\newtheorem{remark}{Remark}

\PassOptionsToPackage{table}{xcolor}
\newcolumntype{C}{>{\centering\arraybackslash}X}

\newcolumntype{\vsep}{!{\color{black!25}\vrule}}
\newcommand{\code}[1]{\texttt{\hyphenchar\font=`\-\relax #1}}
\newcommand\blfootnote[1]{%
  \begingroup
  \renewcommand\thefootnote{}\footnote{#1}%
  \addtocounter{footnote}{-1}%
  \endgroup
}

\lstdefinestyle{snippet}{
  language=Python,
  basicstyle=\ttfamily\scriptsize,
  keywordstyle=\bfseries,
  commentstyle=\itshape\color{gray},
  columns=fullflexible,
  keepspaces=true,
  frame=single,
  aboveskip=0pt,
  belowskip=0pt,
}
\usepackage[T1]{fontenc}

\title{PACE: Policy-Native Adaptive Decision Timing for Long-Horizon Reasoning}

\author{Chen Li, Zhantao Yang, Fangyi Chen, Han Zhang, Anudeepsekhar Bolimera \& Marios Savvides \\
Carnegie Mellon University \\
\texttt{\{chenli4, zhantaoy, fangyic, hanz3, abolimer, marioss\}@andrew.cmu.edu}
}

\begin{document}

\blfootnote{%
  Project Page: \textcolor{magenta}{\nolinkurl{https://stellar-neuron.github.io/PACE/}}%
}

\maketitle

\vspace{-1em}
\begin{abstract}
\vspace{-0.5em}

Long-horizon reasoning requires deciding not only \emph{what} actions to take, but \emph{how many} to execute open-loop before replanning. This number---the \emph{execution depth}---balances replanning cost against compounding execution errors. 
Most current systems either fix the execution depth as a hand-tuned scalar or adjust it at inference time with heuristic rules decoupled
from the policy; we argue both can be suboptimal.
In this work, we treat the execution depth as a learnable, history-conditioned variable of the policy itself, and propose PACE, a \emph{model-native} VLM policy that jointly predicts what to execute and for how long under a hard decision budget. We evaluate PACE on four long-horizon environments spanning fully and partially observable settings and visual and textual modalities: Sliding Puzzle, Sokoban, ALFWorld, and ScienceWorld. Across all settings, PACE improves success rate by 3.1 $\sim$ 15.6 percentage points, while using fewer decisions on average, achieving empirical success--decision Pareto dominance.

\end{abstract}

\section{Introduction}
\label{sec:intro}
\vspace{-0.5em}
\begin{figure}[H]
    \centering
    \vspace{-0.5em}
    \includegraphics[width=1\linewidth]{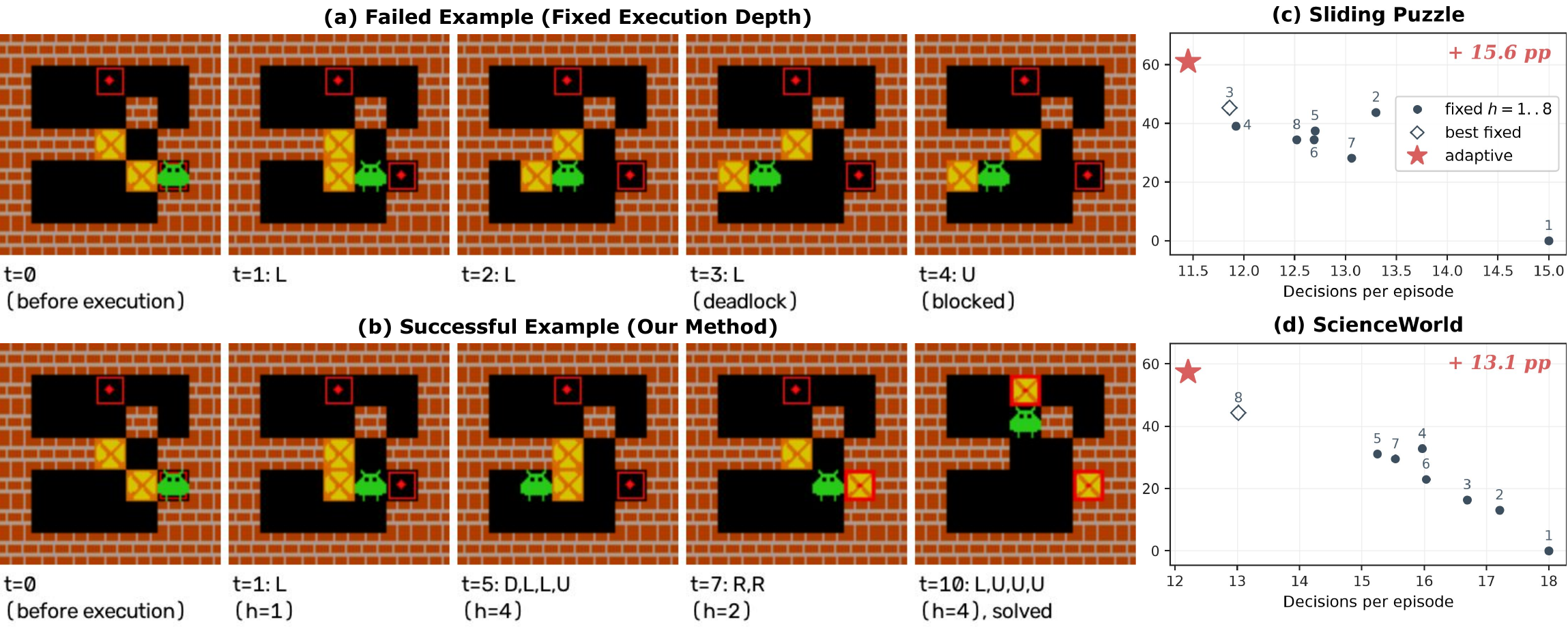}
    \caption{\textbf{Fixed execution depth versus adaptive execution depth on Sokoban game.} The player needs to push the two boxes onto the goal using four actions: left (L), right (R), up (U), down (D).
    (a, b) depict failed and successful examples on the same Sokoban game.
    (a) shows failed fixed execution depth ($h=4$) execution on Sokoban. The model decides \texttt{L,L,L,U} without seeing any intermediate state. Its internal simulation miscounts the pushes, so the third \texttt{L} drives the lower box into a corner at $t=3$, an irreversible deadlock, and the final \texttt{U} walks into a wall.
    (b) shows the same puzzle with adaptive execution depths $h \in \{1,2,4,8\}$. After the single risky push, the agent re-observes ($h=1$). It then executes longer depth of actions through safe moves and solves the puzzle in 4 decisions, within the 6-decision budget.
    (c, d) depict success rate versus mean decisions per episode on Sliding Puzzle and ScienceWorld. The adaptive policy (red star) Pareto-dominates every fixed horizon $h \in \{1,\dots,8\}$: it reaches higher success with fewer decisions (+15.6 and +13.1\,pp over the best fixed $h$).}
    \label{fig:teaser}
\end{figure}

Foundation-models are increasingly asked to solve long-horizon
tasks, such as multi-step puzzles, household manipulation, and
multi-stage scientific procedures, yet they persistently struggle as
the horizon grows.
These tasks require an agent to repeatedly decide what to do, execute
the actions, observe the outcome, and decide again.

Classical work on temporal abstraction
\citep{sutton1999between,bacon2017option} addresses long horizons with
closed-loop \emph{options} whose termination conditions are
task-semantic. Modern long-horizon systems \citep{zhao2023learningfinegrainedbimanualmanipulation,chi2023diffusion,kim2025finetuningvisionlanguageactionmodelsoptimizing,intelligence2025pi05visionlanguageactionmodelopenworld} take a simpler route: a
plan-then-act loop, in which the policy generates a chunk of
actions, executes it \emph{open-loop}---without re-observing the
environment---and then re-observes and replans. We call the number of
primitive actions executed between two consecutive replans the
\emph{execution depth} $h$.
The execution depth induces a trade-off: a short $h$ spends more
replans, each a full backbone forward pass over a fresh observation,
while a long $h$ accumulates compounding error as the imagined and
realized states diverge.

Because every backbone invocation is expensive, the number of replans
per episode is a genuine resource. We make it explicit as a hard
\emph{decision budget} $K$: each replan costs one unit regardless of
its execution depth, and the episode ends once $K$ decisions are
exhausted. Under the same budget, we evaluate Pareto improvements
in success versus actual decision counts and,
separately, success versus environment steps. The central question is then how to spend this budget: at each
decision, how many actions should be executed before looking again? In short, \emph{when to replan}?

We formulate this as a budget-constrained decision problem: the
objective is task success; the decision variable at decision $k$ is
the pair $(h_k, \mathbf{a}_k)$ of an execution depth and an action
chunk of that length; and the constraint is that at most $K$
decisions are made per episode.
Existing systems largely sidestep this problem by fixing $h$ as a single hand-tuned scalar per task. Such as, action chunking in
robot learning \citep{zhao2023learningfinegrainedbimanualmanipulation,chi2023diffusion},  modern
vision--language--action models \citep{kim2025finetuningvisionlanguageactionmodelsoptimizing,intelligence2025pi05visionlanguageactionmodelopenworld}, step-bounded
reasoning in coding, web, and embodied VLM systems
\citep{yao2023react,yang2024sweagentagentcomputerinterfacesenable,koh2024visualwebarenaevaluatingmultimodalagents}.
This is the suboptimality we argue against. In this work, we empirically show that on long-horizon visual reasoning tasks, the optimal $h$ should be state-, task-, and budget-dependent, and a scalar cannot track a target that varies across states.

We propose \textbf{PACE}, a policy-native method with adaptive decision timing that enhances long-horizon reasoning. A lightweight depth head and a compact action
decoder share a single 7B vision--language backbone
(Fig.~\ref{fig:method_pipeline}). The action decoder predicts an action sequence,
while the depth head selects how many actions to
execute before observing and planning again. The two heads are trained jointly:
first an SFT warm start on counterfactual macro-action data, then a
GRPO-style RL objective
\citep{shao2024deepseekmathpushinglimitsmathematical} under the hard
budget $K$. The rollout procedure enforces the budget: episodes
that use all $K$ decisions without completing the task
receive zero task-success reward.
RL therefore trains the policy to succeed within this
limit, without providing $K$ or a remaining-budget
counter as input. To support controlled measurement, we also
release a vision--language puzzle environment with an
exact-solver-grounded macro-action data pipeline
(App.~\ref{app:tasks}).

We evaluate PACE across four environments and seven
settings, covering visual puzzles and text-based
embodied tasks.
Using the same trained model and decision budget,
PACE improves success rate by 3.1--15.6 percentage points
over the best-performing fixed depth while using
fewer decisions in every setting, achieving
success--decision Pareto dominance.
On Sokoban, it also reduces environment steps by
25.5\% (30.1 vs.\ 40.4 per episode).
The success rate advantage persists at budgets different
from training, without retraining
(\S\ref{sec:experiments}).

Our contributions are summarized in three folds:
\begin{itemize}[leftmargin=1.5em,topsep=1pt,itemsep=1pt]

    \item \textbf{(C1) Problem formulation.}
    We formulate execution-depth selection as a decision process
    under a hard budget, with actions executed open-loop between
    observations.
    With the action-generation mechanism fixed, we establish
    a sufficient condition for a history-conditioned depth
    policy to strictly outperform the best fixed depth in
    the same admissible set
    (\S\ref{sec:problem}, Prop.~\ref{prop:improvement};
    proof in App.~\ref{app:proof}).
    \item \textbf{(C2) Unified adaptive policy and infrastructure.}
    We propose a single policy-native VLM with a depth head and a compact
    action head sharing one backbone, trained jointly under one
    objective (\S\ref{sec:method}), along with our self-built vision–language
    puzzle environment and an automatic macro-action data
    pipeline (App.~\ref{app:tasks}).    
    \item \textbf{(C3) Success--decision Pareto dominance.}
    Across seven settings in four environments, PACE
    improves success by 3.1--15.6 percentage points with
    fewer decisions than the best-performing fixed depth
    under the hard budget
    (Table~\ref{tab:main_results}, Fig.~\ref{fig:pareto_decs}).
\end{itemize}

\section{Related Work}
\label{sec:related}

\paragraph{Reinforcement learning for long-horizon agents.}
RL for LLM agents improves credit assignment and
training stability through group-relative optimization
\citep{shao2024deepseekmathpushinglimitsmathematical},
anchor-state grouping
\citep{feng2025groupingrouppolicyoptimizationllm},
milestone-based credit assignment
\citep{wang2026milestoneguidedpolicylearninglonghorizon},
and related methods \citep{ecpo2026,epo2025,empg2025}.
Our focus is orthogonal: PACE learns how many actions
to execute before re-observing under a hard decision
budget.
Integrating these training advances is left to future work.

\paragraph{Adaptive action chunking in robotics and control.}
Fixed-length action chunks are used in imitation
learning \citep{zhao2023learningfinegrainedbimanualmanipulation}
and TD learning \citep{li2025qchunking}.
Subsequent work adjusts execution through inference-time
rules \citep{bid2025,rtc2025,
liang2026adaptiveactionchunkinginferencetime,kwts2026,
nie2026pacephaseawarechunkexecution},
learned values \citep{ach2026,
gireesh2026adaptiveqchunkingofflinetoonlinereinforcement},
or an RL-trained horizon head over a frozen action policy
\citep{zhao2026dynamicexecutionhorizonprediction}.
PACE focuses on allocating a hard decision budget:
it jointly learns actions and execution depth, balancing
extended execution against opportunities for feedback
across fully and partially observable environments.

\paragraph{Adaptive execution in foundation-model agents.}
Concurrent work SPACE \citep{yang2026actmoredecideless}
combines multi-action execution with skill distillation
and execution-time controls.
HPL \citep{hpl2026} uses multi-step groups as preference
supervision units, while early-exit methods
\citep{earlyexit2025} route decisions between models.
PACE jointly learns actions and execution depth through
compact heads within a unified policy under a hard
decision budget, without skill distillation, skill libraries,
a separate preference-optimization objective,
or mid-execution guards.

\paragraph{Temporal abstraction and costly observation.}
Options establish temporal abstraction and policy improvement
through interruption \citep{sutton1999between};
option-critic learns termination \citep{bacon2017option},
while deliberation costs penalize frequent decisions
\citep{harb2018deliberation}.
Costly-observation methods study when to acquire feedback
\citep{nam2021acno,krale2023actthenmeasure},
and adaptive computation allocates reasoning effort
\citep{graves2016act,banino2021pondernet,
yang2026thinkfastslowsteplevel}.
PACE learns how many actions to execute between observations,
studying the trade-off between decision cost and task success
under a hard budget.
Proposition~\ref{prop:improvement} specializes one-step
policy improvement to this budgeted open-loop setting
(App.~\ref{app:proof}).

\section{Method}
\label{sec:method}

PACE jointly learns actions and execution depth
under a hard decision budget.

\subsection{Problem Formulation}
\label{sec:problem}

\paragraph{Environments and observability.} An episodic environment is a tuple $(\mathcal{S},\mathcal{A},P,\Omega,O,r)$; when $O$ reveals the state it is a Markov decision process (MDP), otherwise a partially observable Markov decision process (POMDP) \citep{kaelbling1998pomdp}. Our puzzles have deterministic dynamics and fully observable
states, whereas ALFWorld and ScienceWorld are partially
observable: new observations can provide information about
previously unobserved aspects of the environment.

\paragraph{The budget-$K$ decision process.}
A decision is one planning round, including the initial
plan and subsequent replanning rounds.
Given context $x$---the history of observations and
executed actions---the policy $\pi=(\pi^a,\pi^h)$
outputs an action chunk and an execution depth
$h\in\mathcal{H}=\{1,2,4,8\}$.
The first $h$ actions execute open-loop before the
next observation and decision, unless the episode terminates.
Each decision consumes one unit of a hard budget $K$,
regardless of depth; exhausting $K$ before completion
is budget failure.
Decision count therefore measures planning rounds,
rather than wall-clock time or total inference compute.
The objective is success probability $J_K(\pi)$;
a fixed-depth policy $\pi^{(h)}$ uses a constant depth
with the same action mechanism.
The budget is omitted from policy inputs;
formal definitions appear in App.~\ref{app:proof}.

\paragraph{Why learn the depth?}
\begin{proposition}[When adaptive depth improves success]
\label{prop:improvement}
Fix the action-generation mechanism and a fixed-depth policy $\pi^{(h)}$. Let $Q^{\pi}(x,d)$ denote the success probability when selecting depth $d$ at decision history $x$ and following $\pi$ thereafter, with all budget and termination rules enforced. If a nonterminal history $x^{*}$ is reached with probability $\rho>0$ under $\pi^{(h)}$ and some $h'\in\mathcal{H}$ satisfies $\Delta=Q^{\pi^{(h)}}(x^{*},h')-Q^{\pi^{(h)}}(x^{*},h)>0$, then changing depth only at $x^{*}$ yields a policy $\pi'$ with
\[
J_K(\pi')-J_K(\pi^{(h)})\;=\;\rho\,\Delta\;>\;0.
\]
\end{proposition}

If this condition holds for a best fixed-depth policy
in $\mathcal{H}$, the constructed policy strictly
outperforms every fixed depth in that set under the
same action-generation mechanism
(Cor.~\ref{cor:dominance}).
The proof specializes standard one-step policy
improvement to budgeted open-loop execution.
It shares the policy-improvement principle of option
interruption \citep{sutton1999between}, but selects
execution depth at decision boundaries rather than
revising termination during execution.
Proof and scope appear in App.~\ref{app:proof}.

\subsection{Automated Macro-Action Data Pipeline}
\label{sec:pipeline}

We use exact-solver optimal paths for the puzzles and
built-in task-completing demonstrations, without
optimality guarantees, for ALFWorld and ScienceWorld
(App.~\ref{app:tasks}).
For each reference trajectory $(a_1, \ldots, a_L)$, every starting position $u$ yields
examples $\bigl(x_u, (a_u, \ldots, a_{u+m-1}), m\bigr)$ for $m \in \mathcal{H} = \{1, 2, 4, 8\}$
with $u + m - 1 \le L$, where $x_u$ is the context before action $a_u$ and $m$ is the
target action-chunk length.
Each chunk is predicted from $x_u$ without
intermediate observations, matching open-loop execution.

\subsection{Model Architecture}
\label{sec:arch}

\begin{figure}[t]
    \centering
    \includegraphics[width=0.95\linewidth]{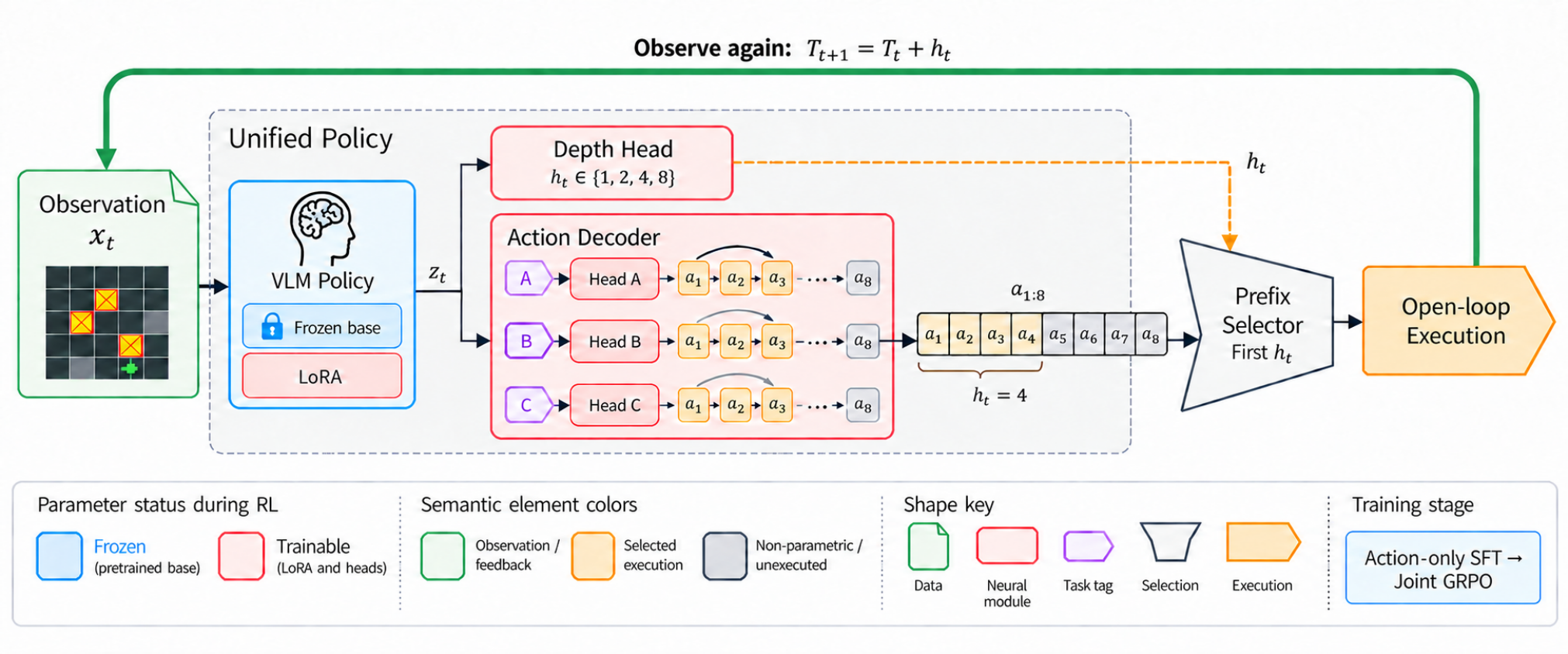}
    \vspace{-1.em}
    \caption{\textbf{PACE architecture.}
    The backbone representation $z_t$ feeds a depth head
    and an autoregressive action decoder.
    The decoder generates eight actions independently
    of the selected depth $h_t$; the first $h_t$ execute
    without intermediate feedback.
    RL jointly updates backbone LoRA and both heads.}
  \label{fig:method_pipeline}
\vspace{-1.25em}
\end{figure}

As shown in Fig.~\ref{fig:method_pipeline}, PACE uses
a shared Qwen2.5-VL-7B-Instruct backbone with a
lightweight depth head and a compact action decoder.
At decision $t$, the backbone encodes the interaction
context into $z_t$, which both branches reuse.
PACE predicts depth and actions directly, without
generating or supervising explicit chain-of-thought text.Decoder specifications appear in
App.~\ref{app:exp_details}.

\paragraph{Factorized depth and action prediction.}

The depth head predicts a distribution over
$\mathcal{H}=\{1,2,4,8\}$.
The action decoder predicts primitive actions from a domain-specific vocabulary
$\mathcal{A}_e$
(Table~\ref{tab:domain_matrix}; vocabulary sizes
in Table~\ref{tab:vocab}), conditioned on $z_t$
and previously generated actions.
Depth selects the executed prefix and is not a
decoder input:
$\pi_\theta(h_t,a_t^{1:h_t}\mid x_t)
=\pi_\theta^h(h_t\mid z_t)
\prod_{j=1}^{h_t}
\pi_\theta^a(a_t^j\mid z_t,a_t^{<j})$.
The decoder generates up to $h_{\max}=8$ actions,
of which the first $h_t$ are executed open-loop.

\paragraph{Compact output parameterization.}

Table~\ref{tab:parameterization} compares output
parameterizations.
A flat classifier requires $\sum_{h \in \mathcal{H}} |\mathcal{A}_e|^h$ depth--chunk
classes, whereas our factorization uses $|\mathcal{H}|$ depth logits and $|\mathcal{A}_e|$
action logits per decoded position.
It composes known actions into chunks without
enumerating complete chunks as separate classes.
Each action output identifies a direction or a
complete command, avoiding token-by-token command
generation and free-text parsing.

\begin{table}[b]
\centering
\footnotesize 
\caption{\textbf{Output parameterization per decision.}
Counts refer to output logits, not total computation.
$T(h)$ is the number of text tokens used to serialize
depth and $h$ commands; $V$ is the LM token vocabulary,
and $\mathcal{A}_e$ is the domain-specific action vocabulary.
PACE selects depth from $\mathcal{H}=\{1,2,4,8\}$,
decodes $h_{\max}=8$ actions, and executes the
selected prefix of length $h$.}
\label{tab:parameterization}
\begin{tabularx}{\linewidth}{@{}lXl@{}}
\toprule
Scheme & Prediction target & Output logits \\
\midrule
Autoregressive LM
& Depth and commands as text
& $T(h)|V|$ \\
Flat joint classifier
& One class per depth--sequence pair
& $\sum_{d\in\mathcal{H}}|\mathcal{A}_e|^d$ \\
PACE
& Depth and $h_{\max}$ autoregressive actions
& $|\mathcal{H}|+h_{\max}|\mathcal{A}_e|$ \\
\bottomrule
\end{tabularx}
\end{table}

\subsection{Training Protocol}
\label{sec:training}

Both SFT and RL adapt the shared backbone through LoRA,
keeping its pretrained weights frozen.
LoRA configurations and training hyperparameters
are provided in App.~\ref{app:exp_details}.

\paragraph{Action-supervised warm start.}
SFT updates the backbone LoRA and action decoder
using teacher-forced cross-entropy on the multi-length
examples from \S\ref{sec:pipeline}, while keeping
the depth head frozen.
Depth selection is learned during RL.
We train three SFT models: one shared by the two puzzles,
one across the five ALFWorld task types, and one across
all ScienceWorld horizon groups.

\paragraph{Joint reinforcement learning.}
GRPO jointly updates the backbone LoRA, action decoder,
and depth head under a hard decision budget.
Each puzzle is trained separately; ALFWorld shares one
model across five task types, and ScienceWorld shares
one across horizon groups with group-specific budgets.
Candidates are compared within each decision point using
the rewards discussed in \S\ref{sec:exp-setup}.
Policy inputs omit budget and step counters.
During training, rollouts stop after at most $K$
decisions; unfinished episodes receive zero
task-success reward, while any progress rewards
remain as specified in \S\ref{sec:exp-setup}.
These budget-constrained outcomes provide the RL
signal for learning action and depth choices that
complete tasks within the available decisions.
In both text environments, only the first action is
masked to admissible commands; subsequent actions execute
without intermediate feedback or updated masks.
Nonterminal invalid actions do not interrupt execution.

\section{Experiments}
\label{sec:experiments}

\subsection{Setup and Evaluation Protocol}
\label{sec:exp-setup}

\paragraph{Tasks and benchmarks.}
We evaluate four environments in seven settings:
Sliding Puzzle, Sokoban, ALFWorld seen/unseen
\citep{shridhar2021alfworldaligningtextembodied},
and ScienceWorld short/medium/long
\citep{wang2022scienceworldagentsmarter5th}.
Our visual puzzle benchmarks pair observations with
exact-solver diagnostics; Sliding Puzzle is reversible,
whereas Sokoban permits irreversible deadlocks.
The text environments provide partially observable
interaction.
Domain properties and construction details appear
in App.~\ref{app:tasks}
(Table~\ref{tab:domain_matrix}).

\paragraph{Task selection and horizon coverage.}
We retain 12 ScienceWorld physics and chemistry task
types based on vocabulary size and reference-trajectory
compatibility with the 150-step evaluation cap.
Short, medium, and long groups use separate budgets
to examine timing across its broad horizon range;
these groups describe reference length, not calibrated
difficulty.
Other environments evaluate mixed-length instances
under a common budget.
ALFWorld retains five families, excluding Pick2,
whose post-SFT success remains below 10\% even under
a relaxed budget.
Integrating complementary credit-assignment advances
\citep{feng2025groupingrouppolicyoptimizationllm,
wang2026milestoneguidedpolicylearninglonghorizon}
to improve Pick2 is outside this study's compute scope.
Selection and grouping details appear in
App.~\ref{app:sciworld_tasks}
and App.~\ref{app:alfworld_tasks}.

\paragraph{Data splits and vocabulary coverage.}
Puzzles use fresh evaluation instances from the
training generators; ALFWorld uses official seen/unseen
splits; ScienceWorld uses disjoint training and evaluation
(task, variation) pairs within each horizon group.
Text-domain vocabularies use training data only.
All evaluation instances remain in the main results,
including those with out-of-vocabulary commands in
their reference paths, and are scored by actual rollouts
(App.~\ref{app:oov}).

\paragraph{Reward settings.}
All domains use sparse success, supplemented by
solver-derived progress in the puzzles and official-score
changes in ScienceWorld.
ALFWorld uses sparse success alone.
We add no further task-specific shaping, testing
both progress-augmented and sparse-only rewards.

\paragraph{Decision budgets.}
Main training and evaluation budgets match:
$K=15$ for Sliding Puzzle, $6$ for Sokoban, $9$ for
ALFWorld, and $6/9/18$ for ScienceWorld
short/medium/long.
Reference lengths inform these choices to impose
decision pressure while permitting longer depths
to extend action capacity to at most $Kh$.
All timing policies share the budget within a setting;
constraint severity need not match across settings.
Budget generalization increases evaluation budgets
without retraining
(\S\ref{sec:budget_transfer};
Table~\ref{tab:decision_budgets}).

\paragraph{Implementation and compute.}
PACE uses Qwen2.5-VL-7B-Instruct
\citep{bai2025qwen25vltechnicalreport}.
Training takes 10--48 hours per domain on eight
NVIDIA A100 80\,GB GPUs.
Training configurations and compute appear in
App.~\ref{app:hyperparameters};
additional zero-shot model configurations appear
in App.~\ref{app:zero_shot_protocol}.

\paragraph{Baselines and depth coverage.}
Fixed-depth and heuristic policies share PACE's
RL-trained checkpoint.
Fixed policies execute prefixes of length
$h\in\{1,\ldots,8\}$ from the eight-action decoder,
directly testing depths outside
$\mathcal{H}=\{1,2,4,8\}$.
We select the highest-success fixed depth in hindsight
per setting, retaining ties.
SFT-only instead reports the best fixed depth
over the same range using the pre-RL checkpoint,
without its untrained depth head.

Heuristics include Repeat-last, Random-$h$
(uniform over $\mathcal{H}$), and cumulative-entropy
and per-action entropy-threshold rules inspired
by AAC~\citep{liang2026adaptiveactionchunkinginferencetime}.
The best-$\tau$ threshold is selected on the evaluation
set per setting, favoring this baseline within its
tested family. More details are discussed in App.~\ref{app:heuristic_baselines}.
We also evaluate the zero-shot backbone.

\paragraph{Metrics and Pareto dominance.}
We report success rate $\bar{y}$ and mean executed
decisions $\bar{D}$ and environment steps $\bar{S}$
over all episodes, including failures.
For $c\in\{\bar{D},\bar{S}\}$, policy $\pi_1$
Pareto-dominates $\pi_2$ when
$\bar{y}_{\pi_1}\ge\bar{y}_{\pi_2}$ and
$c_{\pi_1}\le c_{\pi_2}$, with at least one strict
inequality.
Success--decision and success--interaction dominance
are assessed separately under matched protocols.

\subsection{Main Results}
\label{sec:exp-main}

\begin{table}[t]
\centering
\caption{\textbf{Success rates (\%) across seven settings.}
PACE achieves the highest observed success in every setting.
Joint-RL fixed-depth and heuristic timing policies share
PACE's trained checkpoint; SFT-only uses the checkpoint
before RL.
For each checkpoint and setting, best fixed-$h$ is selected
in hindsight over $h\in\{1,\ldots,8\}$.
$\Delta$ compares PACE with the joint-RL best fixed depth.
$^\dagger$The best-$\tau$ threshold is selected separately
for each setting on the evaluation set, providing an
optimistic comparison within the tested threshold family.}
\label{tab:main_results}
\begin{adjustbox}{max width=\textwidth}
\begin{tabular}{lccccccc}
\toprule
 & & & \multicolumn{2}{c}{\textbf{ALFWorld}}
     & \multicolumn{3}{c}{\textbf{ScienceWorld}} \\
\cmidrule(lr){4-5}
\cmidrule(lr){6-8}
\textbf{Method} & \textbf{Sliding Puzzle} & \textbf{Sokoban}
 & Seen & Unseen & Short & Medium & Long \\
\midrule
Qwen2.5-VL-7B~\citep{bai2025qwen25vltechnicalreport}
 & 0.0 & 0.0 & 5.2 & 3.4 & 0.0 & 1.9 & 1.7 \\
SFT-only (best fixed-$h$)
 & 48.4 & 34.4 & 37.9 & 35.0 & 27.5 & 54.5 & 37.7 \\
\midrule
Repeat-last
 & 0.0 & 3.1 & 18.1 & 10.3 & 0.0 & 0.0 & 0.0 \\
Random-$h$
 & 35.9 & 15.6 & 44.0 & 40.2 & 34.1 & 54.5 & 23.6 \\
Entropy (cumulative)
 & 46.9 & 17.2 & 37.9 & 29.9 & 26.1 & 41.8 & 16.4 \\
Entropy (threshold, best-$\tau$)$^\dagger$
 & 37.5 & 29.7 & 44.8 & 35.0 & 34.8 & 47.3 & 4.9 \\
Joint RL (best fixed-$h$)
 & 45.3\,{\scriptsize($h$=3)}
 & 32.8\,{\scriptsize($h$=8)}
 & 45.7\,{\scriptsize($h$=4/7)}
 & 38.5\,{\scriptsize($h$=7/8)}
 & 34.8\,{\scriptsize($h$=6)}
 & 58.2\,{\scriptsize($h$=5/7)}
 & 44.3\,{\scriptsize($h$=8)} \\
\midrule
\textbf{PACE (ours)}
 & \textbf{60.9} & \textbf{35.9}
 & \textbf{54.3} & \textbf{48.7}
 & \textbf{44.9} & \textbf{69.1} & \textbf{57.4} \\
$\Delta$ vs.\ Joint RL (best fixed) (pp)
 & $+15.6$ & $+3.1$ & $+8.6$ & $+10.3$
 & $+10.1$ & $+10.9$ & $+13.1$ \\
\bottomrule
\end{tabular}
\end{adjustbox}
\vspace{-0.5em}
\end{table}

\paragraph{Success across settings.}
PACE achieves the highest observed success across
all seven settings (Table~\ref{tab:main_results}),
improving over the hindsight-best fixed depth by
3.1--15.6 percentage points.
It also outperforms random, repeat-last, and
entropy-based timing rules evaluated with the same
trained checkpoint.
The corresponding success--interaction Pareto plots
appear in App.~\ref{app:efficiency}
(Fig.~\ref{fig:pareto_steps}).
On Sokoban, the modest 3.1-percentage-point success
gain accompanies 25.5\% fewer environment steps
per episode (30.1 versus 40.4) than the best fixed depth. ScienceWorld short has lower observed success than
medium and long despite shorter reference trajectories.
Because the groups contain different instances and use
different budgets, these rates do not isolate the
effect of horizon length.

\begin{figure}[t]
    \centering
    \includegraphics[width=\linewidth]{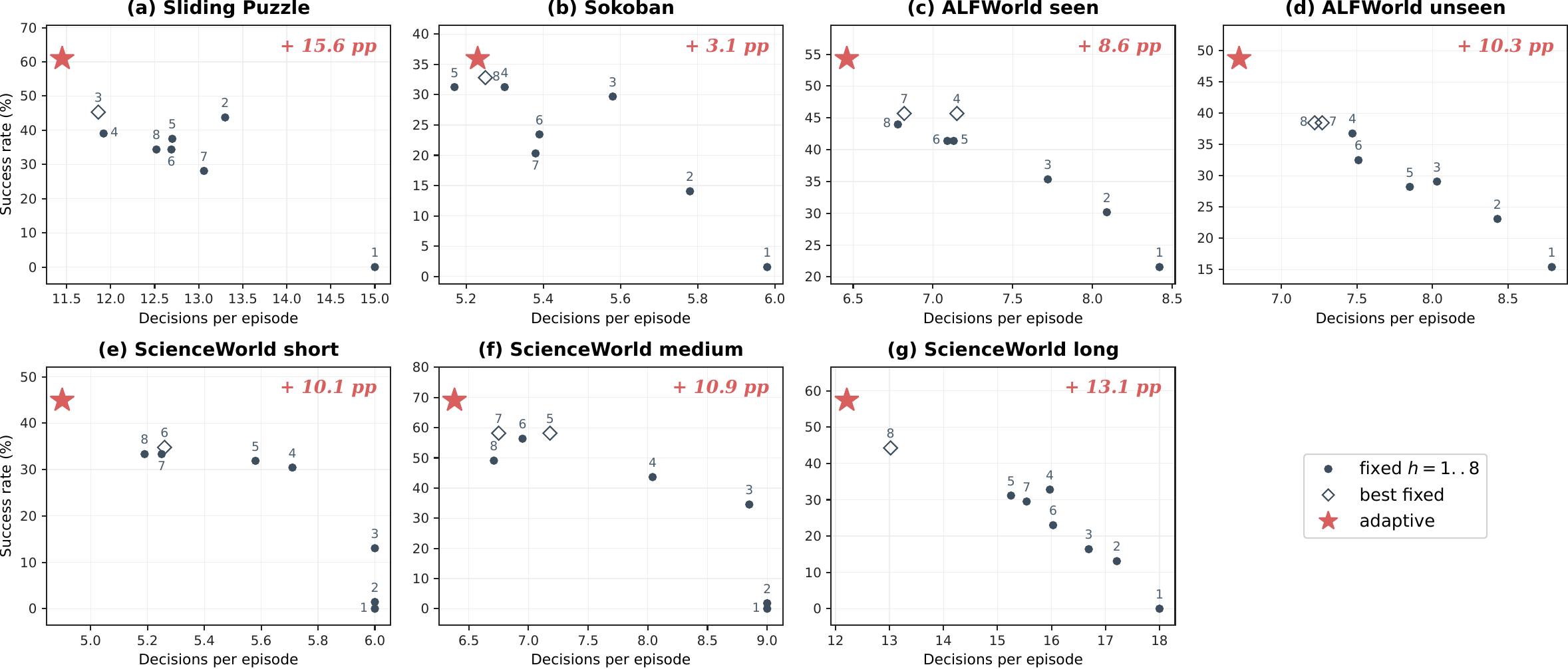}
    \vspace{-0.9em}
    \caption{\textbf{Success--decision trade-offs across
    seven settings.}
    Stars denote PACE; diamonds mark the highest-success
    fixed depths, including ties; circles show the remaining
    tested fixed depths.
    Annotations give success gains over the best fixed depth
    in percentage points.
    Decision counts are averaged over all evaluation
episodes, including failures.
    Axis ranges vary by setting.}
    \vspace{-1em}
    \label{fig:pareto_decs}
\end{figure}

\paragraph{Success--decision trade-off.}
Figure~\ref{fig:pareto_decs} relates success to the
mean number of executed decisions per episode.
Compared with the highest-success fixed-depth policies
on the same RL checkpoint, PACE achieves higher success
with fewer decisions across all seven settings.
This reduces replanning frequency and backbone
context-encoding passes along the executed trajectory,
with replanning cost measured in decision rounds.

\subsection{Ablation Study}
\label{sec:exp-ablation}

\subsubsection{Budget Generalization without Retraining}
\label{sec:exp-budget}
\label{sec:budget_transfer}

Without retraining or changes to the input format,
PACE exceeds the same-checkpoint hindsight-best fixed
depth by 4.7--15.0 percentage points across all seven
larger-budget settings (Table~\ref{tab:budget-transfer}).
The gain reaches 14.5 points on ALFWorld unseen,
showing that the advantage persists on unseen layouts
at a budget different from training.

\begin{table}[t]
\centering
\small
\caption{\textbf{Budget generalization without retraining.}
Each policy is evaluated at a larger decision budget
than used during training, with unchanged policy inputs.
Best fixed-$h$ is selected in hindsight over
$h\in\{1,\ldots,8\}$ at the evaluation budget,
using the same checkpoint as PACE.
Success rates are percentages; $\Delta$ denotes
PACE minus best fixed in percentage points.
}
\label{tab:budget-transfer}
\begin{adjustbox}{max width=\textwidth}
\begin{tabular}{lcccc}
\toprule
Setting
& $K_{\mathrm{train}}\rightarrow K_{\mathrm{eval}}$
& PACE (ours) & Best fixed-$h$ & $\Delta$ (pp) \\
\midrule
Sliding Puzzle
& $5\rightarrow15$ & \textbf{56.2} & 43.8 & $+12.5$ \\
Sokoban
& $3\rightarrow4$ & \textbf{28.1} & 23.4 & $+4.7$ \\
ALFWorld seen
& $5\rightarrow9$ & \textbf{50.9} & 43.1 & $+7.8$ \\
ALFWorld unseen
& $5\rightarrow9$ & \textbf{42.7} & 28.2 & $+14.5$ \\
\midrule
ScienceWorld short
& $6\rightarrow12$ & \textbf{64.2} & 58.2 & $+6.0$ \\
ScienceWorld medium
& $9\rightarrow18$ & \textbf{72.2} & 63.0 & $+9.3$ \\
ScienceWorld long
& $18\rightarrow36$ & \textbf{63.3} & 48.3 & $+15.0$ \\
\bottomrule
\end{tabular}
\end{adjustbox}
\vspace{-1.1em}
\end{table}

\subsubsection{Action-Decoder Freezing versus Joint Optimization}
\label{sec:exp-depthhead}

We compare joint optimization with a variant that
freezes the action decoder at its SFT weights while
updating the depth head and backbone LoRA
(Table~\ref{tab:chunkonly}).
Both variants use adaptive depth.
Because the shared representation remains trainable,
action predictions can still change; this ablation
does not isolate the contribution of timing alone.

Freezing reduces success by 14.1 and 15.6 percentage
points on Sliding Puzzle and Sokoban, respectively,
but improves observed success by 2.6--8.2 points
across all five text-domain settings.
One hypothesis is that the puzzles benefit more from
decoder adaptation, whereas retaining SFT decoder
weights provides useful regularization in the text domains.
We leave controlled experiments investigating the mechanisms underlying these differences
to future work.

\begin{table}[H]
\centering
\small
\caption{\textbf{Depth-only versus joint RL optimization.}
Depth-only updates the depth head and backbone LoRA
while freezing the SFT action decoder;
joint optimization additionally updates the action decoder.
Both use adaptive execution depth.
Success rates are percentages; $\Delta$ is Depth-only
minus Joint in percentage points, computed before rounding.
Bold marks the higher observed success within each setting.}
\label{tab:chunkonly}
\begin{tabular}{lccc}
\toprule
Setting & Depth-only & Joint (PACE) & $\Delta$ (pp) \\
\midrule
Sliding Puzzle
& 46.9 & \textbf{60.9} & $-14.1$ \\
Sokoban
& 20.3 & \textbf{35.9} & $-15.6$ \\
\addlinespace
ALFWorld seen
& \textbf{56.9} & 54.3 & $+2.6$ \\
ALFWorld unseen
& \textbf{53.8} & 48.7 & $+5.1$ \\
\addlinespace
ScienceWorld short
& \textbf{49.3} & 44.9 & $+4.3$ \\
ScienceWorld medium
& \textbf{74.5} & 69.1 & $+5.5$ \\
ScienceWorld long
& \textbf{65.6} & 57.4 & $+8.2$ \\
\bottomrule
\end{tabular}
\vspace{-1em}
\end{table}

\subsection{Zero-Shot Depth Selection}
\label{sec:zero_shot}

We compare prompted self-selected depth with fixed
depths $h\in\{1,2,4,8\}$ within each zero-shot model.
Self-selection does not consistently outperform the
best tested constant: GPT-5.6-Terra achieves
57.8\% versus 43.8\% on Sokoban, but
46.3\% versus 58.2\% on ScienceWorld short.
Thus, prompting variable execution depth alone
does not ensure effective timing.
These results provide supplementary context;
PACE's gains are established through same-checkpoint
comparisons.
Full results and model-specific protocols appear
in App.~\ref{app:zero_shot}.

\section{Efficiency}
\label{sec:efficiency}

\paragraph{Decision and interaction costs.}
Beyond the per-episode trade-offs
(Fig.~\ref{fig:pareto_decs};
App.~\ref{app:efficiency}),
PACE reduces aggregate decisions per success in all
seven settings and steps per success in six relative
to every highest-success fixed-depth comparator,
including costs from failed episodes
(Table~\ref{tab:cost_per_success}).
On ScienceWorld medium, steps per success lie between
the two tied comparators.

\paragraph{Backbone amortization.}
Reusing one backbone representation per decision gives
$D/S=1/\bar{\ell}$ encodings per executed action,
where $D$ and $S$ count decisions and actions,
and $\bar{\ell}=S/D$ is mean executed depth.
This reduces encoding work per action relative to
one-action decisions, with potential latency and
throughput benefits.

\section{Discussion}
\label{sec:discussion}

\begin{figure}[t!]
    \centering
    \includegraphics[width=0.825\linewidth]{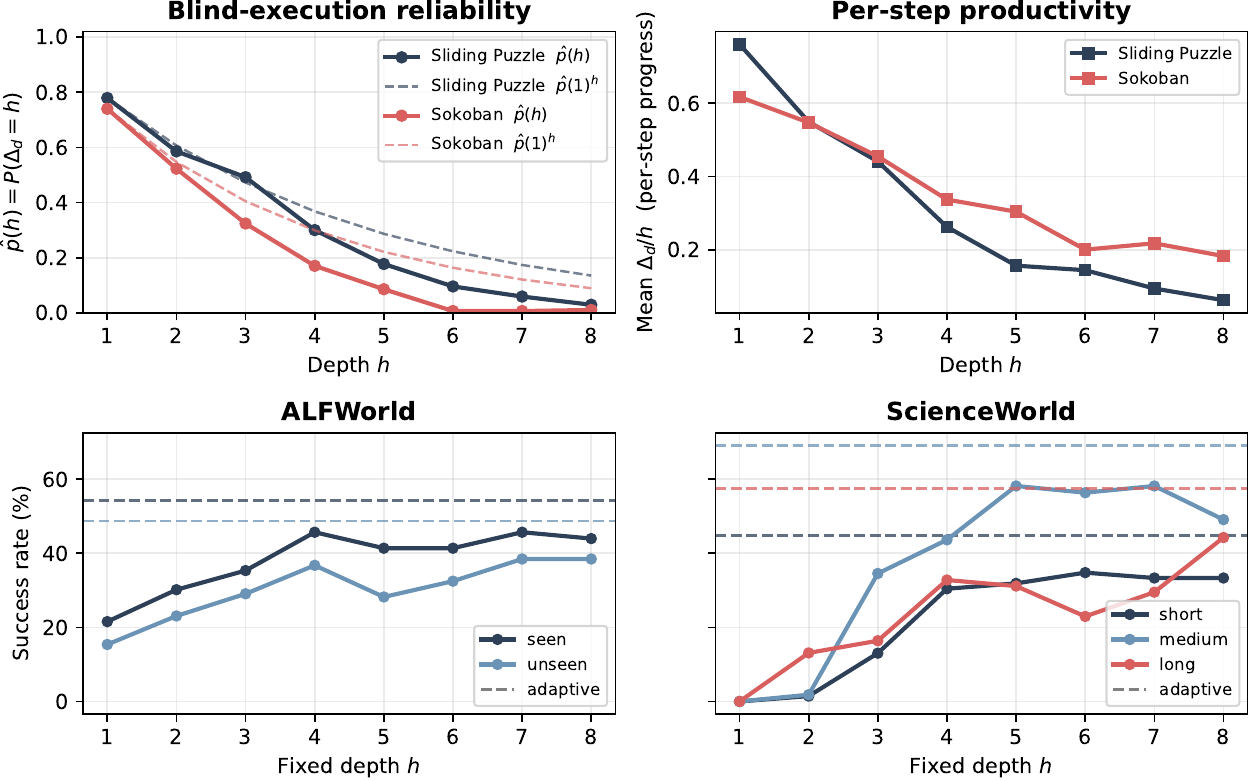}
    \vspace{-1em}
    \caption{\textbf{Execution reliability and depth--success frontiers.}
    Top: puzzle reliability $\hat{p}(h)$ versus $\hat{p}(1)^h$
    and per-step solver progress.
    Bottom: fixed-depth versus PACE success in ALFWorld
    and ScienceWorld using the same checkpoint and
    decision budget within each setting.}
    \label{fig:open_loop_and_depth_frontier}
    \vspace{-1em}
\end{figure}

\paragraph{Execution reliability versus decision cost.}
Longer execution saves decisions but reduces puzzle
optimal-path reliability below $\hat{p}(1)^h$ at larger
depths, with generally lower per-step solver progress
(Fig.~\ref{fig:open_loop_and_depth_frontier}).
Shallow execution instead limits action capacity
under the budget.
In text domains, no tested fixed depth matches
PACE's success on the same checkpoint.

\begin{figure}[H]
    \centering
    \includegraphics[width=0.9\linewidth]{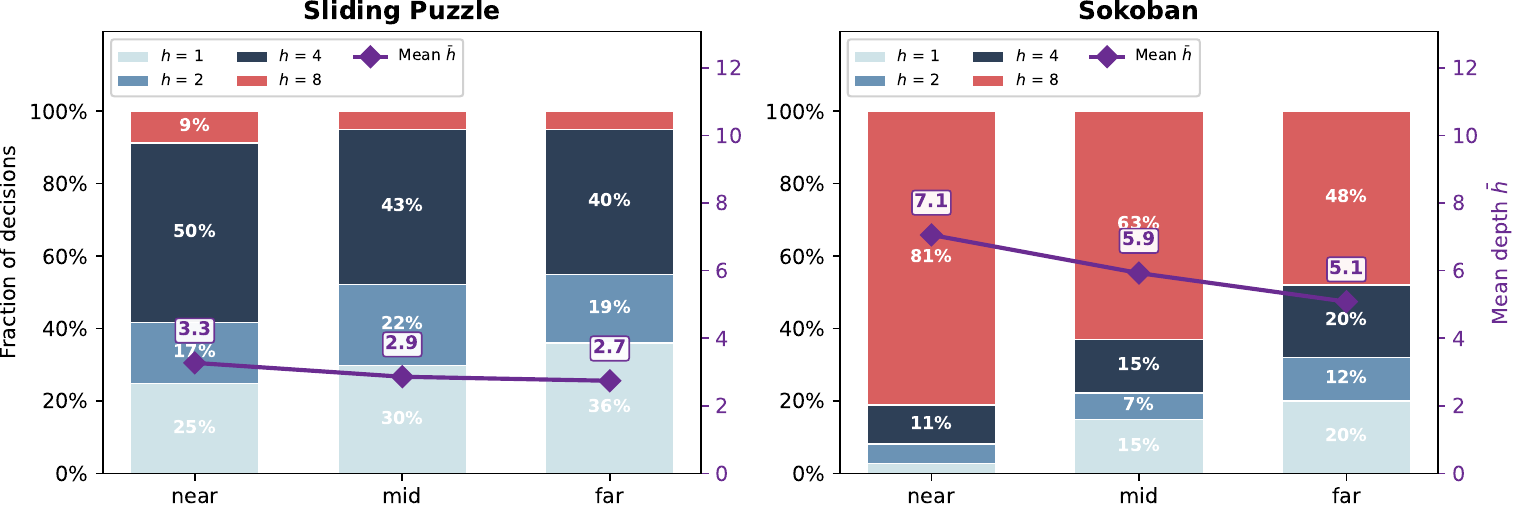}
    \vspace{-1em}
    \caption{\textbf{State-conditioned depth under matched
    training and evaluation budgets.}
    Bars show selected-depth distributions by remaining
    solver distance; purple lines show mean depth.
    Both puzzles exhibit a near-deep/far-shallow pattern.}
    \vspace{-1em}
    \label{fig:h_frac_samek}
\end{figure}

\paragraph{State-conditioned depth allocation.}
In both puzzles, mean selected depth decreases
from near-goal to far-from-goal states under matched
budgets (Fig.~\ref{fig:h_frac_samek}) and relaxed
evaluation budgets (App.~\ref{app:depth_allocation}).
PACE uses all four depths in every setting
(Fig.~\ref{fig:depth_distribution}).
On ScienceWorld medium, it selects $h=8$ in 82.9\%
of decisions yet exceeds fixed-8 success by 20
percentage points
(Fig.~\ref{fig:open_loop_and_depth_frontier}),
showing that a dominant depth preference can coexist
with adaptive gains.
These patterns do not identify which individual
depth choices cause the improvement. Training dynamics and failure cases in App.~\ref{app:training_stability}.

\section{Conclusion}

We introduced PACE, which jointly learns actions
and execution depth under a hard decision budget.
Across four environments and seven settings, it
improves success over the same-checkpoint best fixed
depth while using fewer decisions, with gains
persisting under budget generalization.
State-conditioned depth patterns further characterize
the learned policy.
These results support policy-native decision timing:
learning what actions to take and when to reason again.

\bibliography{iclr2027_conference}
\bibliographystyle{iclr2027_conference}

\appendix

\section{Notation}
\label{app:notation}

Table~\ref{tab:notation} summarizes the notation used in
our formulation, architecture, training, and evaluation.
The theoretical history includes decision boundaries for
budget accounting; policy inputs omit explicit budget and
step counters. Selected depth and mean executed depth are
distinct quantities.

\begingroup
\small
\setlength{\tabcolsep}{5pt}
\renewcommand{\arraystretch}{1.12}
\begin{longtable}{@{}p{0.24\textwidth}p{0.72\textwidth}@{}}
\caption{Notation. Index meanings are local to their context;
probabilities and percentage-point differences use distinct
units as specified below.}
\label{tab:notation}\\
\toprule
Symbol & Meaning \\
\midrule
\endfirsthead

\multicolumn{2}{l}{\tablename~\thetable\ (continued)}\\
\toprule
Symbol & Meaning \\
\midrule
\endhead

\midrule
\multicolumn{2}{r}{Continued on next page}\\
\endfoot

\bottomrule
\endlastfoot

\multicolumn{2}{l}{\textbf{Environment and decision protocol}}\\*
$\mathcal{S},\ s$
& Environment state space and an individual state. \\

$\mathcal{A},\ a$
& Primitive action space and an individual primitive action. \\

$P(s'\mid s,a)$
& Environment transition kernel. \\

$\Omega,\ O$
& Observation space and state-to-observation mapping. \\

$r$
& Binary terminal task-success reward in the problem
formulation; distinct from shaped training rewards. \\

$x,\ x_t$
& Decision history, and the history at decision $t$.
In the proofs, it includes observations, executed actions,
and past decision boundaries. \\

$T_t$ & Environment step at which decision $t$ begins; $T_{t+1} = T_t + h_t$ unless the episode terminates earlier. \\

$K$
& Hard decision budget: each policy invocation costs
one unit, irrespective of execution depth. \\

$K_{\mathrm{train}},\ K_{\mathrm{eval}}$
& Decision budgets enforced during training and evaluation. \\

$\mathcal{H}$
& Learned depth-selection set, $\{1,2,4,8\}$.
Main fixed-depth baselines additionally test $3,5,6,7$. \\

$h,\ h_t,\ h_{\max}$
& Selected execution depth, its value at decision $t$,
and the maximum decoded length ($h_{\max}=8$).
Termination can shorten the executed prefix. \\

$\bar{h}$ & Mean selected depth over a set of decision points (Figs.~\ref{fig:h_frac} and~\ref{fig:h_frac_samek}); distinct from the mean executed depth $\bar{\ell}$. \\

\addlinespace
\multicolumn{2}{l}{\textbf{Policy and architecture}}\\*
$\pi=(\pi_a,\pi_h)$
& Policy comprising the action-generation mechanism
and the depth-selection rule. \\

$\theta,\ \pi_\theta$
& Parameters and joint policy of the learned model. \\

$z_t$
& Backbone representation of the available policy
context at decision $t$, reused by both output branches. \\

$\pi_\theta^h,\ \pi_\theta^a$
& Parameterized depth distribution and
action distribution, respectively. The selected depth
is not an input to the action decoder. \\

$a_t^j,\ a_t^{<j},\ a_t^{1:h_t}$
& Action at position $j$ within decision $t$, preceding
action prefix, and the first $h_t$ predicted actions. \\

$\mathcal{A}_e$
& Fixed decoder action vocabulary for domain $e$;
its entries represent complete primitive actions. \\

$\mathcal{A}_{\mathrm{adm}}(s)$
& Actions admissible at state $s$. In the text domains,
first-action masking uses
$\mathcal{A}_{\mathrm{adm}}(s)\cap\mathcal{A}_e$. \\

\addlinespace
\multicolumn{2}{l}{\textbf{Budgeted success and improvement result}}\\*
$J_K(\pi)$
& Episode success probability under policy $\pi$ and
budget $K$, with the task distribution and other limits fixed. \\

$\pi^{(h)},\ f_h$
& Fixed-depth policy using the shared action mechanism;
$f_h$ is its shorthand in the evaluation tables. \\

$Q^\pi(x,d)$
& Conditional success probability after choosing depth
$d$ at history $x$, using the shared action mechanism,
and following $\pi$ thereafter. \\

$x^*,\ \rho$
& Nonterminal history at which a deviation is considered,
and its positive arrival probability under $\pi^{(h)}$. \\

$h',\ \pi'$
& Alternative depth and the policy that uses it only at
$x^*$, otherwise following $\pi^{(h)}$. \\

$\Delta$ (Proposition)
& Continuation-value gap
$Q^{\pi^{(h)}}(x^*,h')-Q^{\pi^{(h)}}(x^*,h)$,
measured in probability units. \\

$\Pi_{\mathcal{H}},\ h^\star$
& Full-history depth-policy class over $\mathcal{H}$
with the action mechanism fixed, and a depth maximizing
$J_K(\pi^{(h)})$ over $h\in\mathcal{H}$. \\

\addlinespace
\multicolumn{2}{l}{\textbf{Supervision and training rewards}}\\*
$L$ & Length of a reference trajectory $(a_1, \ldots, a_L)$. \\

$x_u, a_u, m$ (SFT)
& Reference context, reference action at trajectory
position $u$, and supervised depth
$d\in\mathcal{H}$; $d$ is not a depth-head label. \\

$G$
& GRPO group size during RL training. \\

\addlinespace
\multicolumn{2}{l}{\textbf{Diagnostics and efficiency}}\\*
$\delta$ & Solver-defined distance to the goal in the puzzles. \\

$\hat p(h)$
& Empirical solver-defined optimal-path reliability
of a length-$h$ open-loop action chunk in the puzzles. \\

$\hat p(1)^h$
& Independent-step reference constructed from
one-step empirical reliability. \\

$\Delta d,\ \mathbb{E}[\Delta d/h]$
& Reduction in solver distance to the goal and the
length-normalized progress statistic used in
Fig.~\ref{fig:open_loop_and_depth_frontier}. Here $d$ denotes solver distance. \\

$D_i,\ S_i,\ y_i$
& Executed decisions, executed environment steps,
and binary success for evaluation episode $i$.
Step counts include invalid attempts. \\

$D,\ S$
& Decision and environment-step counts over the same
specified trajectory or collection of trajectories. \\

$\bar{\ell}$
& Mean executed depth across decisions:
$\sum_i S_i/\sum_i D_i$.
Its reciprocal is encodings per executed action under
one backbone encoding per decision. \\

$C_{\mathrm{dec}},\ C_{\mathrm{step}}$
& Aggregate executed cost per observed success:
$\sum_i D_i/\sum_i y_i$ and
$\sum_i S_i/\sum_i y_i$, respectively.
Both include failed-episode costs. \\

$c$ & Generic cost in Pareto comparisons, $c \in \{\bar{D}, \bar{S}\}$. \\

$\pi_1, \pi_2$ & Generic policies in the Pareto-dominance definition. \\

$\tau$
& Threshold used by an entropy-based timing rule. \\

$\Delta$ (results tables)
& Success-rate difference in percentage points;
its direction is specified by the table. \\

$n$ (puzzle geometry)
& Side length of an $n\times n$ Sliding Puzzle board;
unrelated to the evaluation sample size. \\

\end{longtable}
\endgroup

\section{Proofs}
\label{app:proof}

\subsection{Setup}

An episodic environment is a tuple $(\mathcal{S},\mathcal{A},P,\Omega,O,r)$ as in \S\ref{sec:problem}. A \emph{decision history} $x$ is the full interaction record at a decision boundary: all observations and executed primitive actions so far, including the placement of past decision boundaries and the number of decisions taken. Histories therefore encode the budget accounting: with hard budget $K$, the number of remaining decisions is a function of $x$. All comparisons below fix the initial task distribution, the decision budget $K$, environment step limits, and termination rules.

\begin{definition}[Budget-$K$ decision process]
\label{def:process}
At each decision boundary with history $x$, a policy $\pi=(\pi_a,\pi_h)$ selects an execution depth $h\in\mathcal{H}$ via $\pi_h(\cdot\,|\,x)$ and an action chunk via $\pi_a$; the first $h$ actions are executed open-loop, yielding the next decision history. Every decision consumes one unit of the budget regardless of $h$. An episode ends in success (terminal reward $1$), in environment-side termination, or in \emph{budget failure} when $K$ decisions are exhausted before success. The objective is the success probability $J_K(\pi)$. For $h\in\mathcal{H}$, the \emph{fixed-depth arm} $\pi^{(h)}$ selects depth $h$ at every decision, with the same action mechanism $\pi_a$.
\end{definition}

For a policy $\pi$ and non-terminal decision history $x$, $Q^{\pi}(x,d)$ is as in Proposition~\ref{prop:improvement}:
\[
Q^{\pi}(x,d)\;=\;\Pr\bigl[\text{episode succeeds}\;\big|\;x;\ \text{depth }d\text{ now, actions from }\pi_a,\ \pi\text{ thereafter}\bigr],
\]
where the conditional law includes the budget accounting of Definition~\ref{def:process}, environment step limits, and termination rules.

\begin{assumption}[Shared action mechanism]
\label{ass:shared}
At history $x$, the action chunk is sampled from the fixed kernel $\pi_a(\cdot\,|\,x)$; the selected depth determines the executed prefix and is not an input to the action decoder (\S\ref{sec:arch}). All policies compared below share this kernel and differ only in depth selection.
\end{assumption}

\subsection{Proof of Proposition~\ref{prop:improvement}}

\begin{proof}
A decision history contains its entire prefix, so $x^{*}$ occurs at most once along any trajectory, and $\pi'$ is well defined. The two policies agree at every history other than $x^{*}$; in particular they agree on the event that $x^{*}$ is never reached, and they induce the same distribution over trajectory prefixes up to and including arrival at $x^{*}$, so the arrival probability is $\rho$ under both. Conditioning on arrival, the success probability is $Q^{\pi^{(h)}}(x^{*},h)$ under $\pi^{(h)}$ and $Q^{\pi^{(h)}}(x^{*},h')$ under $\pi'$: the deviation is taken once and both policies thereafter coincide with $\pi^{(h)}$, and any change in the downstream visitation distribution induced by the deviation is part of the conditional law defining $Q^{\pi^{(h)}}(x^{*},h')$, not an omission from it. The law of total probability over the arrival event gives the identity, and positivity follows from $\rho>0$ and the premise.
\end{proof}

\begin{corollary}[Dominance over fixed depths]
\label{cor:dominance}
Let $\Pi_{\mathcal{H}}$ be the class of history-conditioned depth policies over $\mathcal{H}$ under Assumption~\ref{ass:shared}, and let $h^{\star}\in\arg\max_{h\in\mathcal{H}}J_K(\pi^{(h)})$ (the maximum exists; $\mathcal{H}$ is finite). Then
\[
\sup_{\pi\in\Pi_{\mathcal{H}}}J_K(\pi)\;\geq\;\max_{h\in\mathcal{H}}J_K(\pi^{(h)}),
\]
since every fixed arm belongs to $\Pi_{\mathcal{H}}$. If, in addition, the premise of Proposition~\ref{prop:improvement} holds for a best fixed-depth arm $\pi^{(h^{\star})}$, the constructed deviation policy $\pi'\in\Pi_{\mathcal{H}}$ satisfies
\[
J_K(\pi')\;>\;\max_{h\in\mathcal{H}}J_K(\pi^{(h)}) :
\]
a history-conditioned depth policy strictly outperforms every fixed depth in $\mathcal{H}$.
\end{corollary}

\subsection{Scope and interpretation}

\begin{remark}[Scope of the depth set]
\label{rem:scope}
The guarantee is internal to $\mathcal{H}=\{1,2,4,8\}$: weak dominance follows from class inclusion over the \emph{same} depth set, and the strict statement compares against the best arm in $\mathcal{H}$. Our experiments additionally evaluate fixed depths $3,5,6,7$; comparisons against these are empirical results outside the class covered by Corollary~\ref{cor:dominance}.
\end{remark}

\begin{remark}[Existence versus realizability]
The result establishes the existence of an improving
history-conditioned depth policy under the stated condition.
It does not guarantee that the chosen policy parameterization
can represent this policy or that training will recover it;
the performance of the learned policy is assessed empirically.
\end{remark}

\begin{remark}[Relation to existing guarantees]
\label{rem:lineage}
The proof specializes a standard one-step policy-improvement argument to budgeted open-loop execution. It shares the principle underlying option interruption \citep{sutton1999between}, but selects depth at the decision boundary rather than changing an ongoing option's termination rule, and uses no intermediate observations during execution. AQC studies related adaptive-length dominance under discounted return with advantage-based depth selection \citep{gireesh2026adaptiveqchunkingofflinetoonlinereinforcement}; timing-as-an-action has also been formalized explicitly \citep{pmlr-v238-zhou24c}. Our result gives a sufficient condition for improvement in the budget-$K$ success objective over full-history depth policies, without claiming a new policy-improvement principle or a guarantee for the learned network.
\end{remark}

\section{Task illustrations and instance generation}
\label{app:tasks}

\begin{table}[H]
\centering
\small
\caption{\textbf{Domain coverage and training reward settings.}}
\label{tab:domain_matrix}

\begingroup
\setlength{\tabcolsep}{4pt}
\renewcommand{\arraystretch}{1.2}

\begin{tabularx}{\linewidth}{
    @{}
    >{\raggedright\arraybackslash}p{0.18\linewidth}
    *{4}{>{\centering\arraybackslash}X}
    @{}
}
\toprule
& \multicolumn{2}{c}{\textbf{Visual environments}}
& \multicolumn{2}{c}{\textbf{Text environments}} \\
\cmidrule(lr){2-3}
\cmidrule(lr){4-5}
\textbf{Property}
& \textbf{Sliding Puzzle}
& \textbf{Sokoban}
& \textbf{ALFWorld}
& \textbf{ScienceWorld} \\
\midrule

Observability
& Full
& Full
& Partial
& Partial \\

Reversibility
& Reversible
& \makecell{Irreversible pushes;\\deadlocks}
& Task-dependent
& \makecell{Irreversible\\outcomes possible} \\

Action vocabulary
& Four directions
& Four directions
& 5,698 commands
& 658 commands \\

Solver diagnostics
& Available
& Available
& Not used
& Not used \\

Horizon coverage
& Mixed horizons
& Mixed horizons
& Mixed horizons
& \makecell{Short / medium\\ / long} \\

Reward setting
& \makecell{Solver progress\\+ sparse success}
& \makecell{Solver progress\\+ sparse success}
& Sparse success only
& \makecell{Official-score change\\+ sparse success} \\

\bottomrule
\end{tabularx}
\endgroup

\vspace{4pt}
\begin{minipage}{\linewidth}
\footnotesize
\textit{Notes.}
ALFWorld's vocabulary is the union of admissible commands
over the training set; ScienceWorld's is built from the
training split of the 12 selected tasks.
Solver diagnostics measure open-loop action quality.
\end{minipage}
\end{table}

\begin{figure}[t]
    \centering
    \includegraphics[width=0.9\linewidth]{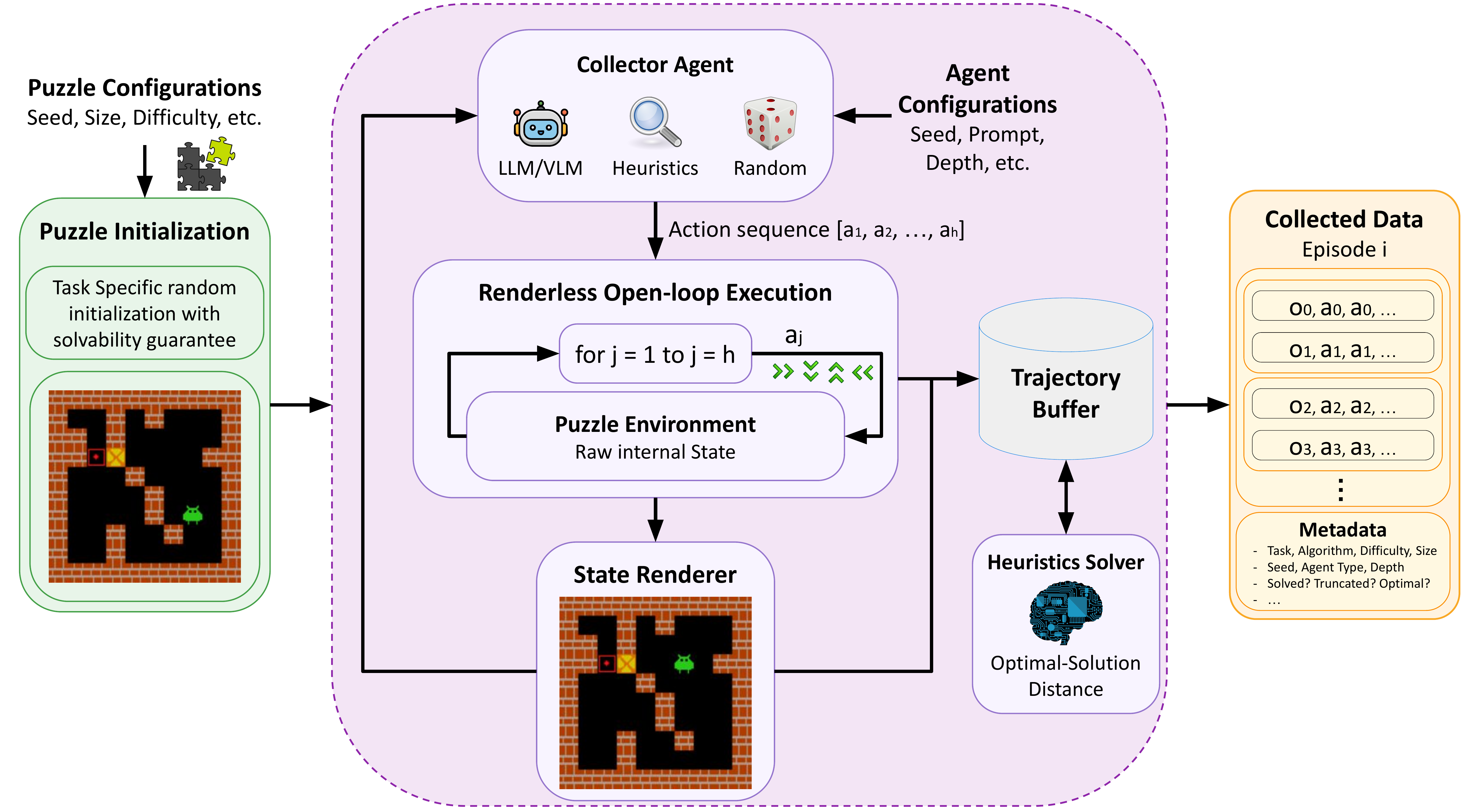}
    \caption{\textbf{Detailed pipeline for data generation and collection.} For each episode, a reproducible and configurable random environment and a selected collector agent are initialized. The agent predicts a chunk for open-loop execution and renders the next state for prediction through a separate renderer. All chunks and states are saved and compared against the heuristics solver for ground-truth-based signal.}
    \label{fig:data_env}
\end{figure}

\begin{figure}[ht]
    \centering
    \includegraphics[width=0.95\linewidth]{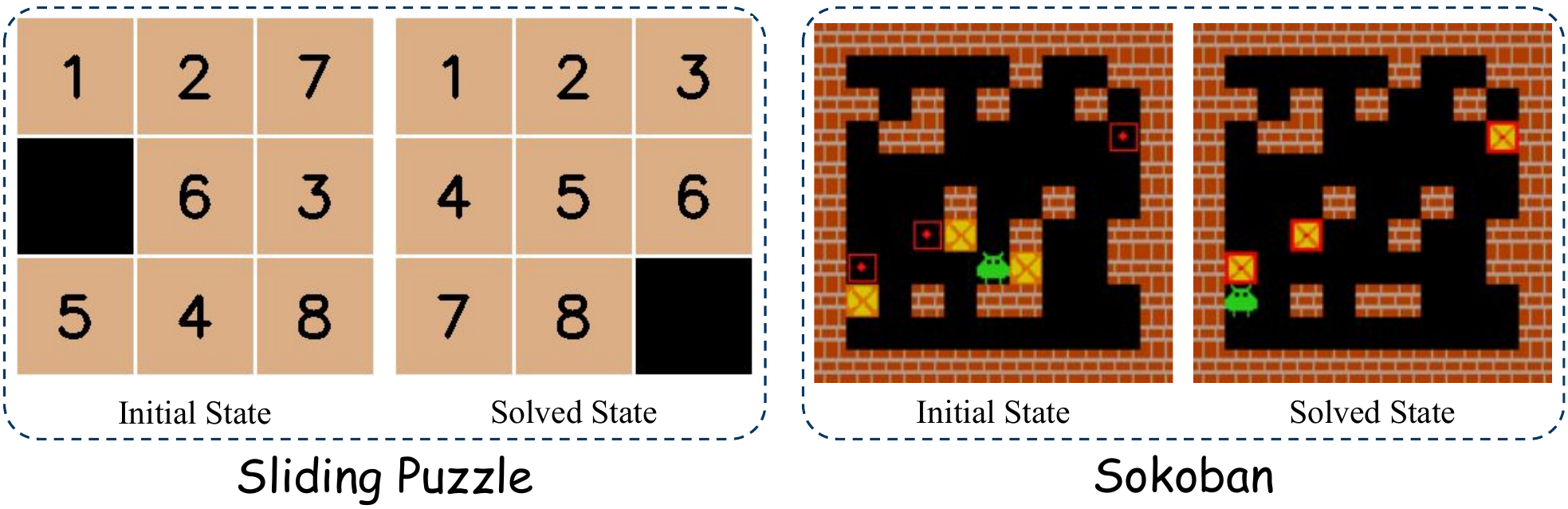}
    \caption{\textbf{Task illustrations.}
    For each task we show one representative instance, with the
    \emph{initial state} on the left and the \emph{goal state} on the
    right of each pair.
    \emph{Left pair:} Sliding Puzzle ($3{\times}3$).  Tiles must be
    rearranged into canonical row-major order (goal) by sliding tiles
    one at a time into the empty cell; every primitive action is
    reversible.
    \emph{Right pair:} Sokoban.  The agent (\textsc{p}layer) must push
    every box (\textsc{b}) onto a goal cell ($\bullet$); pushes are
    not reversible by the agent, so a wrongly emitted long chunk can
    render the instance permanently unsolvable.
    Both tasks require long-horizon planning under known dynamics, but differ in reversibility: Sliding Puzzle moves are reversible, whereas Sokoban pushes can
create irreversible deadlocks.}
    \label{fig:task_examples}
\end{figure}

We show the four task families used throughout our experiments: Sliding Puzzle, Sokoban, ScienceWorld, and ALFWorld.
Fig.~\ref{fig:task_examples} shows the start and goal states of one
representative instance from each image based task family, Fig.~\ref{fig:collection_snippets} shows the simplified codes for each text based task family. Each task family covers different properties and attributes, detailed in Table~\ref{tab:domain_matrix} We also show the data collection and generation engine and our environment pipeline in Figure~\ref{fig:data_env}

\begin{figure}[t]
\centering
\begin{minipage}[t]{0.49\linewidth}
\begin{lstlisting}[style=snippet, title={\textbf{ScienceWorld}}]
env = ScienceWorldEnv(envStepLimit=10**6)
env.load(task, variation, simplificationStr="")
templates = env.get_possible_actions()
obs, info = env.reset()
msgs = [system_prompt(env.taskdescription(),
                      templates)]

for _ in range(K):
    msgs.append(user(render_state(obs, info)))
    reply = model(msgs, schema)
    msgs.append(reply)
    for a in resolve(reply):
        obs, r, done, info = env.step(a)
        if done: break
    if done: break

success = info["score"] == 100
\end{lstlisting}
\end{minipage}\hfill
\begin{minipage}[t]{0.49\linewidth}
\begin{lstlisting}[style=snippet, title={\textbf{ALFWorld}}]
infos = textworld.EnvInfos(won=True,
            admissible_commands=True)
env_id = textworld.gym.register_games(
    [game_file], infos, batch_size=1,
    max_episode_steps=150,
    wrappers=[AlfredDemangler(), AlfredInfos])
env = textworld.gym.make(env_id)
(obs,), info = env.reset()
msgs = [system_prompt(task_of(obs), templates)]

for _ in range(K):
    msgs.append(user(render_state(obs, info)))
    reply = model(msgs, schema)
    msgs.append(reply)
    for a in resolve(reply):
        (obs,),_,(done,),info = env.step([a])
        if done: break
    if done: break

success = info["won"][0]
\end{lstlisting}
\end{minipage}
\caption{\textbf{Simplified data collection loop on the official
ScienceWorld and ALFWorld packages.} At each decision point the model
returns a chunk, which is executed open-loop; only the state after
its last action is rendered into the next user turn.}
\label{fig:collection_snippets}
\end{figure}

\subsection{Sliding Puzzle}

\subsubsection{Mechanics.}
A Sliding Puzzle instance is a square grid of size $n \times n$ with
$n^2 - 1$ numbered tiles and one empty cell.
At each primitive timestep the policy selects an action from $\{\text{up},
\text{down}, \text{left}, \text{right}\}$, sliding a tile from the
indicated direction into the empty cell.
A solved instance is one whose tile configuration matches a fixed
target (canonical row-major: tiles $1, 2, \ldots, n^2-1$ in
left-to-right, top-to-bottom order, with the empty cell in the
bottom-right).

\subsubsection{Reversibility.}
Every primitive action is reversible by the inverse direction;
configurations are reachable iff their permutation has even parity
(half of the $(n^2)!$ configurations).
This rules out unrecoverable states from primitive errors: a wrongly
emitted long chunk only \emph{wastes} primitive actions, never causes
the instance to become impossible.

\subsubsection{Difficulty parameter.}
We characterize instance difficulty by the optimal
solution length: the minimum number of primitive actions
needed to reach the goal, computed by an exact A$^*$
solver with the linear-conflict heuristic.
Longer optimal solutions require more actions,
but do not necessarily imply greater difficulty
for every policy.

\subsubsection{Instance generation.}
To the best of our knowledge, there are no existing standardized Sliding Puzzle benchmark that allows generating random setup with designated difficulty, we implement and evaluate on our own Sliding Puzzle environment.
Each instance is generated by random walking from the goal
configuration for the requested number of primitive steps and
verifying via A$^*$ that the shortest path back has the requested
depth. Train and test instances are distinct samples drawn from
the same generator with the same shortest-path-depth filter.

\subsection{Sokoban}

\subsubsection{Mechanics.}
A Sokoban instance is a 2D grid containing walls, free cells, the
agent (\textsc{p}layer), one or more boxes, and one or more goal
locations (one per box).
At each primitive timestep the agent selects from $\{\text{up},
\text{down}, \text{left}, \text{right}\}$.
A move into a free cell or onto a goal cell shifts the agent;
a move into a box that is itself adjacent to a free cell pushes the
box one step in the same direction; a move into a box adjacent to a
wall or another box is no-op.
A solved instance is one in which every box occupies a goal cell.

\subsubsection{Irreversibility.}
Box pushes are not reversible by the agent: there is no \texttt{pull}
action.
Pushing a box into a corner against two walls produces an
unrecoverable deadlock.
This asymmetry is the source of Sokoban's ``planning-dominant''
character: a wrongly emitted long chunk can render the instance
permanently unsolvable, in contrast to Sliding Puzzle where
mistakes are reversible.

\subsubsection{Difficulty parameter.}

We parameterise difficulty by \emph{box count}: instances with more
boxes have larger combinatorial state spaces and longer optimal
plans on average.
Optimal plans are computed by a domain-specific Sokoban solver
(based on iterative deepening with deadlock detection); we filter generated instances to those whose optimal-plan length lies within a range
appropriate for our main and budget-generalization decision budgets (Table~\ref{tab:decision_budgets}).

\subsubsection{Instance generation.}
To the best of our knowledge, there are no existing standardized Sokoban benchmark that allows generating random setup with designated difficulty, we implement and evaluate on our own Sokoban environment.
Each instance is generated through reverse pull scrambling with randomized retries, and walls are additionally greedily added to the map at locations at shuffled cells. Visualization assets are partially borrowed from~\citet{SchraderSokoban2018}.

\begin{table}[t]
\centering
\small
\caption{Evaluated task types in ScienceWorld and ALFWorld.
Selection criteria are detailed in
App.~\ref{app:sciworld_tasks} and
App.~\ref{app:alfworld_tasks}, respectively.}
\label{tab:evaluated_tasks}
\begin{tabularx}{\textwidth}{ll>{\raggedright\arraybackslash}X}
\toprule
\textbf{Benchmark} & \textbf{Category / Family} & \textbf{Tasks / Official Label} \\
\midrule
\multirow{6}{*}{\textbf{ScienceWorld}} 
  & Phase change & \code{boil}, \code{melt}, \code{freeze}, \code{change-the-state-of-matter-of} \\
  \addlinespace[3pt]
  & Measurement  & \code{use-thermometer}, \code{measure-melting-point-known-substance}, \code{measure-melting-point-unknown-substance} \\
  \addlinespace[3pt]
  & Electricity  & \code{power-component}, \code{power-component-renewable-vs-nonrenewable-energy}, \code{test-conductivity}, \code{test-conductivity-of-unknown-substances} \\
  \addlinespace[3pt]
  & Chemistry    & \code{chemistry-mix} \\
\midrule
\multirow{5}{*}{\textbf{ALFWorld}} 
  & Pick \& Place    & \code{pick\_and\_place\_simple} \\
  & Examine in Light & \code{look\_at\_obj\_in\_light} \\
  & Clean \& Place   & \code{pick\_clean\_then\_place\_in\_recep} \\
  & Heat \& Place    & \code{pick\_heat\_then\_place\_in\_recep} \\
  & Cool \& Place    & \code{pick\_cool\_then\_place\_in\_recep} \\
\bottomrule
\end{tabularx}
\end{table}

\subsection{ScienceWorld}

\subsubsection{Mechanics.}
A ScienceWorld~\citep{wang2022scienceworldagentsmarter5th} instance is an elementary
science task set in a text-simulated environment of interconnected
rooms (kitchen, workshop, greenhouse, \ldots) populated with objects,
containers, and devices governed by simple thermodynamic, electrical,
and chemical rules.
An instance is identified by a (task, variation) pair; the variation
fixes the target substance or object, its location, and the room layout.
At each primitive timestep the agent issues a free-form English command
instantiating one of the task's action templates (e.g.\
\texttt{open OBJ}, \texttt{move OBJ to OBJ}, \texttt{connect OBJ to OBJ},
\texttt{focus on OBJ}, \texttt{wait}), with objects named as they appear
in the observations.
A command the parser does not accept leaves the state unchanged but still
consumes a primitive step.
We use the unsimplified environment: doors must be opened and there is
no teleport action.
Each task defines an ordered sequence of subgoals and a score in
$[0, 100]$; a solved instance is one in which every subgoal has been met
(score $100$).

\subsubsection{Irreversibility.}
ScienceWorld is the only environment in our suite in which an episode can
fail before its budget is exhausted: \texttt{focus on} an object other
than the one the task designates ends the episode immediately with
score $-100$.
Many physical transformations (mixing substances, melting or boiling a
sample, using up the only instance of an object) likewise cannot be undone.
A wrongly emitted long chunk can therefore either terminate the
episode outright or remove the object the task depends on.

\subsubsection{Reference horizons.}
We group instances by the length of the simulator's
reference action chunk for each variation.
Observed reference lengths span 6--22 actions in
the short group, 23--32 in medium, and 35--153 in long,
with decision budgets $K=6$, $9$, and $18$, respectively.
These groups describe reference horizons rather than
calibrated difficulty levels.
Larger budgets provide greater action capacity
for the longer-horizon groups.

\subsubsection{Instance generation.}
We use official ScienceWorld package for data collection.
Instances are not generated procedurally: each is a (task, variation)
pair drawn from ScienceWorld evaluation set, split by horizon length.
The three buckets span 12 task types (Table~\ref{tab:evaluated_tasks}).

\subsubsection{ScienceWorld Task Selection}
\label{app:sciworld_tasks}

\paragraph{Three cuts to twelve.}
Selection accounts for grounded vocabulary size and compatibility
with our execution limit.
(i) \emph{Plant/genetics (4 removed).}
In our integration audit, these tasks generate entities at runtime
and contribute approximately 550k of the 599k enumerated commands
(92\%). With a 512-dimensional decoder output, a 600k-way
classification layer alone would require approximately 307M
weights, making this inventory impractical for our training setup.
(ii) \emph{Inclined-plane (3 removed).}
Sampled gold trajectories reach 333 actions, exceeding our
150-step evaluation cap.
We exclude these tasks because their reference trajectories
are not consistently compatible with this cap; shorter
solutions may nevertheless exist.
(iii) \emph{Remaining biology and paint-mixing (11 removed).}
We further restrict the study to the laboratory procedures
in Table~\ref{tab:evaluated_tasks} to control vocabulary growth.
The paint-mixing pair alone adds approximately 14.9k commands.
The decoder vocabulary constructed from the training
split of the 12 retained task types contains 658 commands
(Table~\ref{tab:vocab}).

\paragraph{Scope of evaluation.}
Our factorized decoder avoids enumerating complete action
chunks, while its output layer scales linearly with the
grounded action vocabulary (\S\ref{sec:arch}).
Task selection keeps this inventory manageable under our
architecture and training resources.
Accordingly, our ScienceWorld findings apply to the selected
12 task types under the stated action interface and execution
cap; they do not establish performance on the full 30-task
benchmark.

\paragraph{Horizon stratification.}
Episodes are grouped by reference-trajectory length.
Held-out median lengths are 18, 25, and 57 for the
short, medium, and long groups, respectively.
Reference lengths are used for grouping and budget
configuration, never as policy inputs.

\paragraph{Task list.}
Table~\ref{tab:evaluated_tasks} lists the twelve retained task types
by category.

\subsection{ALFWorld}

\subsubsection{Mechanics.}
An ALFWorld~\citep{shridhar2021alfworldaligningtextembodied} instance is a TextWorld~\citep{cote2019textworldlearningenvironmenttextbased}
game aligned with a household task from the ALFRED
benchmark~\citep{shridhar2020alfredbenchmarkinterpretinggrounded}.
The room contains numbered receptacles (\texttt{drawer 3},
\texttt{countertop 1}, \ldots) and objects placed in or on them; object
locations are hidden until the agent visits, and if necessary opens, the
receptacle holding them.
At each primitive timestep the agent issues a command from a fixed set of
templates, such as \texttt{go to}, \texttt{open}, \texttt{close},
\texttt{take \ldots\ from}, \texttt{move \ldots\ to}, \texttt{heat},
\texttt{cool}, \texttt{clean}, and \texttt{use}.
The agent must go to a receptacle before interacting with it and carries
at most one object at a time.
A command the game does not accept is answered with ``Nothing happens.''
and still consumes a primitive step.
Success is binary with no partial credit: a solved instance is one in
which the task's goal condition holds, e.g.\ a cooled apple placed on the
countertop.

\subsubsection{Reversibility.}
Some action errors are recoverable: misplaced objects
can often be retrieved and moved elsewhere.
However, recovery, unsuccessful searches, and invalid
commands consume environment steps and may prevent
completion within the decision budget.

\subsubsection{Difficulty parameter.}
ALFWorld has no explicit difficulty parameter, and all instances share
the decision budget $K{=}9$, which is enforced but not disclosed to the
model.
We instead report results separately on the two evaluation splits:
\emph{valid-seen}, whose room layouts appear in the ALFWorld training
games, and \emph{valid-unseen}, whose layouts do not.
As a reference solution length we record the length of the plan returned
by ALFWorld's PDDL planner; since the planner has full observability of
object locations, this is a lower bound that an agent which must search
for objects cannot generally attain.

\subsubsection{Instance generation.}
We use official alfworld package for data collection.
Instances are the fixed ALFWorld evaluation games rather than generated
ones: we take every game in the valid-seen and valid-unseen splits
except those of the \texttt{pick\_two\_obj\_and\_place} task type
(Table~\ref{tab:evaluated_tasks}).

\subsubsection{ALFWorld Task Selection}
\label{app:alfworld_tasks}

We evaluate five of the six official task families
under the official seen/unseen splits
(Table~\ref{tab:evaluated_tasks}),
excluding \texttt{pick\_two\_obj\_and\_place} (Pick2)
for the following reasons.

\paragraph{Limited post-SFT performance.}
Pick2 success remains below 10\% after supervised
warm start, even under step-wise control with
$K=150$.
This low starting success motivates prioritizing
timing comparisons on the retained families.
Published prompting baselines also vary substantially:
BEACON reports Pick2 success of 0--3.7\% with
Qwen2.5-1.5B-Instruct and 3.2--23.8\% with
Qwen2.5-7B-Instruct
\citep{wang2026milestoneguidedpolicylearninglonghorizon}.
These provide context rather than matched comparisons
with our model.

\paragraph{Training investment and study scope.}
Prior work demonstrates that Pick2 benefits from
finer-grained credit assignment.
With Qwen2.5-7B-Instruct, BEACON reports Pick2 success
of 64.7\%, 79.2\%, and 90.0\% for GRPO, GiGPO,
and BEACON, respectively
\citep{wang2026milestoneguidedpolicylearninglonghorizon}.
GiGPO's ALFWorld configuration uses 150 iterations
with $16\times8$ trajectories per iteration,
approximately 19,200 training trajectories
\citep{feng2025groupingrouppolicyoptimizationllm}.
This is a benchmark-wide training allocation,
not a Pick2-specific requirement.
These advances are complementary to our focus on
decision timing.
Under our compute budget, we prioritize the retained
families and leave integrating existing training
techniques to improve Pick2 performance to future work.
Our ALFWorld results therefore concern the five
selected families.

\paragraph{Horizon coverage.}
Reference trajectories in the five retained task families
have a median length of 14 actions and a 90th percentile
of approximately 33 actions.

\paragraph{Task list.}
Table~\ref{tab:evaluated_tasks} lists the five retained
ALFWorld task families and their official labels.

\section{More Experiment Details}
\label{app:exp_details}

\subsection{Model architecture details and Training Hyperparameters}
\label{app:hyperparameters}

\subsubsection{Model Architecture}

All four environments, including both puzzles, use
Qwen2.5-VL-7B-Instruct
\citep{bai2025qwen25vltechnicalreport}
with LoRA adapters \citep{hu2021loralowrankadaptationlarge}
in the language backbone.
Pretrained backbone weights, including the vision
encoder, remain frozen.
SFT updates the adapters and action decoder while
freezing the depth head; joint RL updates all three.

The depth head predicts $h_t\in\mathcal{H}=\{1,2,4,8\}$
from the shared representation $z_t$.
The action decoder autoregressively generates eight
actions from $z_t$ and preceding action embeddings,
independently of the selected depth.
Each environment uses an output projection over its
action vocabulary $\mathcal{A}_e$.

\begin{table}[t]
\centering
\small
\setlength{\tabcolsep}{6pt}
\renewcommand{\arraystretch}{1.12}
\caption{\textbf{Architecture and LoRA configuration.}
Settings are shared across all four environments,
except for action-vocabulary size.}
\label{tab:arch_settings}
\begin{tabular}{@{}ll@{}}
\toprule
Parameter & Value \\
\midrule
\multicolumn{2}{@{}l}{\textit{Backbone adaptation}} \\
LoRA rank / alpha & 16 / 32 \\
LoRA dropout & 0.05 \\
LoRA targets & Language-model attention and MLP projections \\
Precision & bf16 \\
\addlinespace
\multicolumn{2}{@{}l}{\textit{Depth head}} \\
Architecture & Linear: $3{,}584 \rightarrow 4$ \\
Depth menu $\mathcal{H}$ & $\{1,2,4,8\}$ \\
\addlinespace
\multicolumn{2}{@{}l}{\textit{Autoregressive action decoder}} \\
Layers / attention heads & 2 / 4 \\
Hidden / FFN dimension & 512 / 1,024 \\
Dropout / activation & 0.1 / GELU \\
Normalization & Pre-norm \\
Context projection & Linear: $3{,}584 \rightarrow 512$ \\
Action embedding dimension & 512 \\
Maximum depth & 8 actions \\
\addlinespace
\multicolumn{2}{@{}l}{\textit{Output vocabulary size $|\mathcal{A}_e|$}} \\
Sliding Puzzle / Sokoban & 4 / 4 \\
ALFWorld / ScienceWorld & 5,698 / 658 \\
\bottomrule
\end{tabular}
\end{table}

\subsubsection{RL Training Hyperparameters}

\begin{table}[t]
\centering
\small
\setlength{\tabcolsep}{4pt}
\renewcommand{\arraystretch}{1.12}
\caption{\textbf{Main-experiment RL hyperparameters.}
Backbone learning rates apply to LoRA adapters.
ScienceWorld budgets correspond to short, medium,
and long episodes.}
\label{tab:rl_settings}
\begin{adjustbox}{max width=\textwidth}
\begin{tabular}{@{}lcccc@{}}
\toprule
Hyperparameter
& Sliding & Sokoban & ALFWorld & ScienceWorld \\
\midrule
Training cycles
& 50--100 & 50--100 & 50--100 & 50--100 \\
Effective batch size
& 128 & 128 & 128 & 128 \\
GRPO group size
& 8 & 8 & 8 & 8 \\
Training budget $K_{\mathrm{train}}$
& 15 & 6 & 9 & 6 / 9 / 18 \\
Depth menu $\mathcal{H}$
& \multicolumn{4}{c}{$\{1,2,4,8\}$} \\
\addlinespace
GRPO clip $\epsilon$
& 0.05 & 0.05 & 0.20 & 0.20 \\
KL coefficient $\beta$
& 0.05 & 0.20 & 0.02 & 0.02 \\
Depth entropy coefficient $\alpha_h$
& 0.01 & 0.01 & 0.01 & 0.01 \\
Action entropy coefficient $\alpha_a$
& 0.001 & 0.001 & 0.001 & 0.003 \\
\addlinespace
Backbone LoRA learning rate
& $5\times10^{-7}$ & $2\times10^{-7}$
& $5\times10^{-6}$ & $5\times10^{-6}$ \\
Head learning rate
& $7\times10^{-6}$ & $3\times10^{-6}$
& $1\times10^{-4}$ & $1\times10^{-4}$ \\
\bottomrule
\end{tabular}
\end{adjustbox}
\end{table}

\subsection{Training Dynamics}
\label{app:training_stability}

\begin{figure}[t]
    \centering
    \includegraphics[width=\linewidth]
    {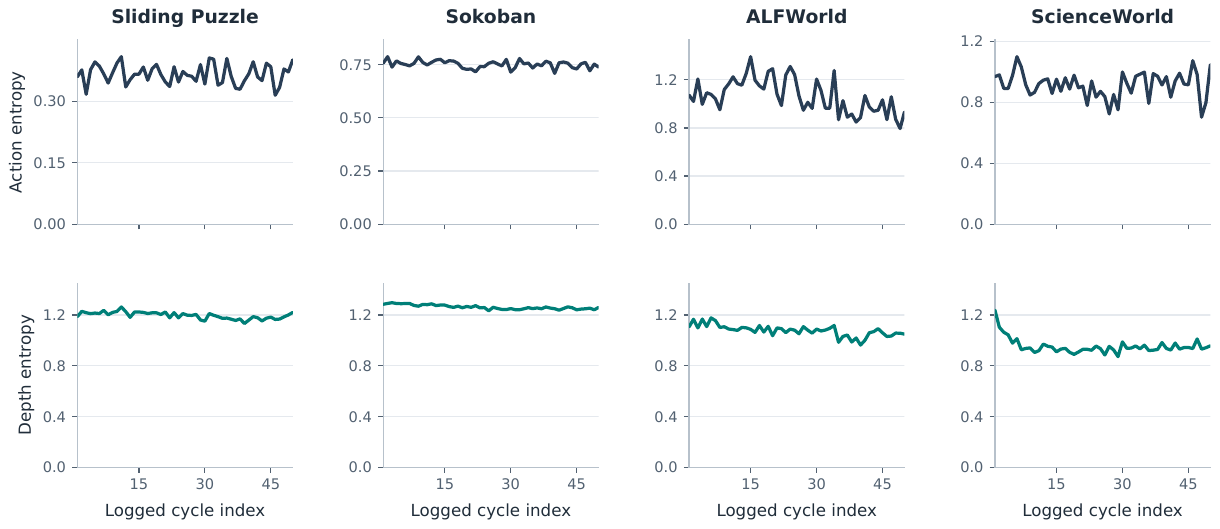}
    \caption{\textbf{Action and depth entropy during RL rollouts.}
    Curves show logged mean action entropy (top) and depth
    entropy (bottom) for the first 50 cycle records in each
    environment, without smoothing.
    The horizontal axis indexes records in file order,
    rather than absolute training-cycle IDs.
    Action-entropy scales vary by environment;
    depth-entropy scales are shared.}
    \label{fig:training_entropy}
\end{figure}

Figure~\ref{fig:training_entropy} tracks action and depth
entropy during RL rollouts.
Depth entropy remains relatively steady in the puzzles,
declines gradually in ALFWorld, and decreases initially
before stabilizing in ScienceWorld.
Neither mean action nor mean depth entropy approaches
zero over the displayed interval.

\subsection{Baselines and Heuristics}
\label{app:heuristic_baselines}

\paragraph{Repeat-last.}
This baseline follows the dynamic action-repetition idea of
FiGAR~\citep{sharma2017figar}, applied at evaluation time to our
trained policy without retraining. At each decision, the depth head
selects $h \in \mathcal{H}$ as usual, but only the first proposed action
is kept and repeated $h$ times open-loop. Everything else (checkpoint,
decision budget, and evaluation episodes) matches the main evaluation.
Unlike FiGAR, which learns the repetition count, we reuse the trained
depth head for this purpose. The baseline therefore keeps PACE's timing
but removes the variety of actions within a chunk, so the gap to
PACE reflects the value of what is executed rather than when the
policy re-observes.

\paragraph{Entropy (threshold, best-$\tau$).}
This baseline shortens execution when the policy is uncertain: the
executed depth is truncated once the predictive entropy exceeds a
threshold $\tau$, so uncertain states lead to earlier re-observation.
Thresholds are set as fractions of the maximum possible entropy, so
they do not depend on a particular checkpoint. For each setting, we
report the best $\tau$ from the tested grid, chosen in hindsight on the
evaluation set (marked $\dagger$ in Table~\ref{tab:main_results}). This
choice favors the baseline, so PACE's advantage over it is a
conservative comparison.

\subsection{Training Details and Decision Budgets}
\label{app:training_details}

\paragraph{Decision-budget configuration.}
Table~\ref{tab:decision_budgets} lists the main and
budget-generalization configurations.
Budget choices were informed by reference-trajectory
lengths to impose decision pressure while allowing
longer execution depths to extend action capacity.
A fixed depth $h$ permits at most $Kh$ primitive
actions, subject to termination and environment limits;
this capacity does not guarantee task completion.
ALFWorld uses one model and a common budget across
its five retained task families.
ScienceWorld uses one model trained on all three
horizon groups, with a separate budget for each group.

\begin{table}[t]
\centering
\small
\caption{\textbf{Hard decision budgets.}
Main experiments match training and evaluation budgets.
Budget-generalization experiments increase the evaluation
budget without retraining.
ALFWorld uses the same five selected task families
in training and evaluation.}
\label{tab:decision_budgets}
\begin{adjustbox}{max width=\textwidth}
\begin{tabular}{lcc}
\toprule
Setting
& Main $K$ (train = eval)
& Generalization (train $\rightarrow$ eval) \\
\midrule
Sliding Puzzle      & 15 & $5 \rightarrow 15$ \\
Sokoban             & 6  & $3 \rightarrow 4$ \\
ALFWorld            & 9  & $5 \rightarrow 9$ \\
ScienceWorld short  & 6  & $6 \rightarrow 12$ \\
ScienceWorld medium & 9  & $9 \rightarrow 18$ \\
ScienceWorld long   & 18 & $18 \rightarrow 36$ \\
\bottomrule
\end{tabular}
\end{adjustbox}
\end{table}

\paragraph{Action vocabularies and output heads.}
Table~\ref{tab:vocab} reports the fixed action vocabulary
$\mathcal{A}_e$ used by each domain's decoder.
The puzzles use four movement directions.
ALFWorld uses 5,698 commands, obtained by taking the union
of admissible commands over the training set.
ScienceWorld uses 658 commands constructed from the
training split of the 12 selected task types.
Each action-decoder output has $|\mathcal{A}_e|$ logits.

\begin{table}[h]
\centering
\small
\caption{\textbf{Per-domain action vocabulary sizes}. The puzzle domains use the four
move directions. The ALFWorld vocabulary is the union of admissible commands
over the training set; the ScienceWorld vocabulary is built from the training
split of the 12 selected tasks.}
\label{tab:vocab}
\begin{tabular}{lr}
\toprule
Setting & $|\mathcal{A}_e|$ \\
\midrule
Sliding Puzzle / Sokoban & 4 \\
ALFWorld & 5{,}698 \\
ScienceWorld (12 tasks) & 658 \\
\bottomrule
\end{tabular}
\end{table}

\paragraph{Checkpoint sharing.}
Table~\ref{tab:model_sharing} summarizes model sharing.
SFT updates the backbone LoRA and action decoder
while keeping the depth head frozen.
RL jointly updates the backbone LoRA, action decoder,
and depth head; pretrained backbone weights remain
frozen throughout both stages.
For ALFWorld, both stages use the same five task types:
Pick \& Place, Clean \& Place, Heat \& Place,
Cool \& Place, and Examine in Light.
Pick2 is excluded from both stages.

\begin{table}[t]
\centering
\small
\caption{Model sharing during supervised and reinforcement
learning.}
\label{tab:model_sharing}
\begin{tabular}{
    p{0.20\linewidth}
    p{0.34\linewidth}
    p{0.34\linewidth}
}
\hline
Domain & SFT & RL \\
\hline
Sliding Puzzle
& Shared with Sokoban
& Separate puzzle-specific model \\

Sokoban
& Shared with Sliding Puzzle
& Separate puzzle-specific model \\

ALFWorld
& One model across five task types
& One model across the same five task types \\

ScienceWorld
& One model across three horizon groups
& One model with mixed-budget training \\
\hline
\end{tabular}
\end{table}

\paragraph{Out-of-vocabulary handling.}
\label{app:oov}
Action vocabularies are constructed from training data only.
Held-out episodes are flagged when their expert demonstration
contains a command outside the vocabulary;
Table~\ref{tab:oov} reports the incidence within the evaluated
task families.
All main-table results use the full evaluation sets:
flagged episodes remain in the denominator and are scored
by their actual rollouts, with no score imputed.

A flag indicates incomplete coverage of the expert path,
not necessarily an unsolvable task.

Across the audited configurations, 9 of the 17 flagged
ALFWorld episodes and 5 of the 13 flagged ScienceWorld
episodes are solved by at least one configuration
through an in-vocabulary trajectory.
Our results assess adaptive timing under the fixed action
inventories and reference-path coverage reported here.
The puzzles use a closed four-action vocabulary and have
no OOV cases.

\begin{table}[h]
\centering
\small
\caption{Held-out episodes whose expert demonstrations contain
out-of-vocabulary commands. Main results retain all episodes,
scored by actual rollouts. Solved counts are aggregated across
audited configurations, not success rates for a single policy.}
\label{tab:oov}
\begin{tabular}{lcc}
\toprule
Split & Flagged fraction & Solved via in-vocab alternatives \\
\midrule
ALFWorld seen          & 3.4\%  & 4/4 \\
ALFWorld unseen        & 11.1\% & 5/13 \\
ScienceWorld held-out  & 7.0\%  & 5/13 \\
\bottomrule
\end{tabular}
\end{table}

\subsection{Efficiency}
\label{app:efficiency}

\begin{figure}[H]
    \centering
    \includegraphics[width=\linewidth]{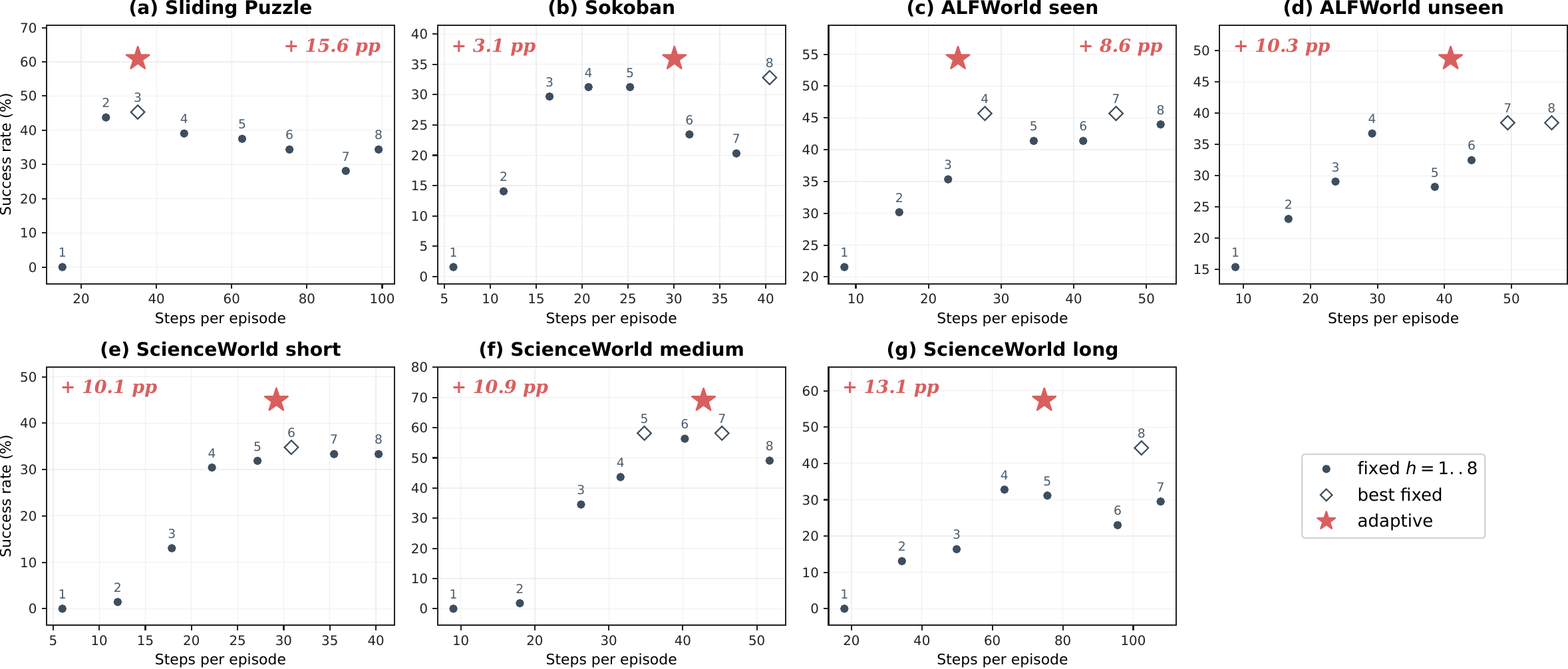}
    \caption{\textbf{Success--interaction trade-offs across seven settings.}
    Stars denote PACE; diamonds mark the highest-success
    fixed depths, including ties; circles show the remaining
    tested fixed depths.
    All fixed-depth policies share PACE's RL-trained checkpoint.
    Annotations give success gains over the best fixed depth
    in percentage points.
    Environment steps are averaged over all evaluation
    episodes, including failures, and include invalid
    action attempts.
    Axis ranges vary by setting.}
    
    \label{fig:pareto_steps}
\end{figure}

\paragraph{Environment interaction steps.}
Figure~\ref{fig:pareto_steps} complements the decision-count
analysis in Fig.~\ref{fig:pareto_decs}.
PACE achieves higher observed success than every tested
fixed depth, while interaction-cost savings vary by setting.
On Sokoban, it uses 30.1 environment steps per episode
versus 40.4 for the highest-success fixed depth.
Decision counts and environment steps therefore capture
distinct aspects of efficiency.

\paragraph{Cost per success.}
We report total executed cost per observed success:
$C_{\mathrm{dec}}=\sum_i D_i/\sum_i y_i$ and
$C_{\mathrm{step}}=\sum_i S_i/\sum_i y_i$,
where $D_i$, $S_i$, and $y_i$ denote episode-level
decision count, environment-step count, and binary success.
Both ratios include costs incurred by failed episodes.
Against the highest-success fixed depths, PACE uses
fewer decisions per success in all seven settings and fewer
steps per success in six (Table~\ref{tab:cost_per_success}).
On ScienceWorld medium, its steps per success lie between
the two tied fixed-depth policies.

\begin{table}[t]
\centering
\small
\caption{\textbf{Aggregate executed cost per success.}
Lower is better.
Fixed-depth comparators are selected by highest success,
not lowest cost per success.
For tied depths, slash-separated values follow the order
in the $h$ column.
Bold indicates that PACE has lower cost than every
listed fixed-depth comparator in that setting.
Costs include both successful and failed episodes.}
\label{tab:cost_per_success}
\begin{adjustbox}{max width=\textwidth}
\begin{tabular}{lccccc}
\toprule
 & & \multicolumn{2}{c}{Decisions per success}
   & \multicolumn{2}{c}{Steps per success} \\
\cmidrule(lr){3-4}
\cmidrule(lr){5-6}
Setting & Fixed $h$
 & Fixed & PACE
 & Fixed & PACE \\
\midrule
Sliding Puzzle
 & 3
 & 26.17 & \textbf{18.79}
 & 77.31 & \textbf{57.54} \\
Sokoban
 & 8
 & 16.00 & \textbf{14.57}
 & 123.19 & \textbf{83.61} \\
ALFWorld seen
 & 4 / 7
 & 15.64 / 14.92 & \textbf{11.89}
 & 60.75 / 100.28 & \textbf{44.30} \\
ALFWorld unseen
 & 7 / 8
 & 18.91 / 18.78 & \textbf{13.79}
 & 128.56 / 145.60 & \textbf{84.00} \\
ScienceWorld short
 & 6
 & 15.13 & \textbf{10.90}
 & 88.62 & \textbf{65.00} \\
ScienceWorld medium
 & 5 / 7
 & 12.34 / 11.59 & \textbf{9.24}
 & 59.81 / 77.84 & 61.95 \\
ScienceWorld long
 & 8
 & 29.41 & \textbf{21.29}
 & 231.19 & \textbf{130.26} \\
\bottomrule
\end{tabular}
\end{adjustbox}
\end{table}

\subsection{Depth Allocation under Relaxed Budgets}
\label{app:depth_allocation}

\begin{figure}[H]
    \centering
    \includegraphics[width=\linewidth]{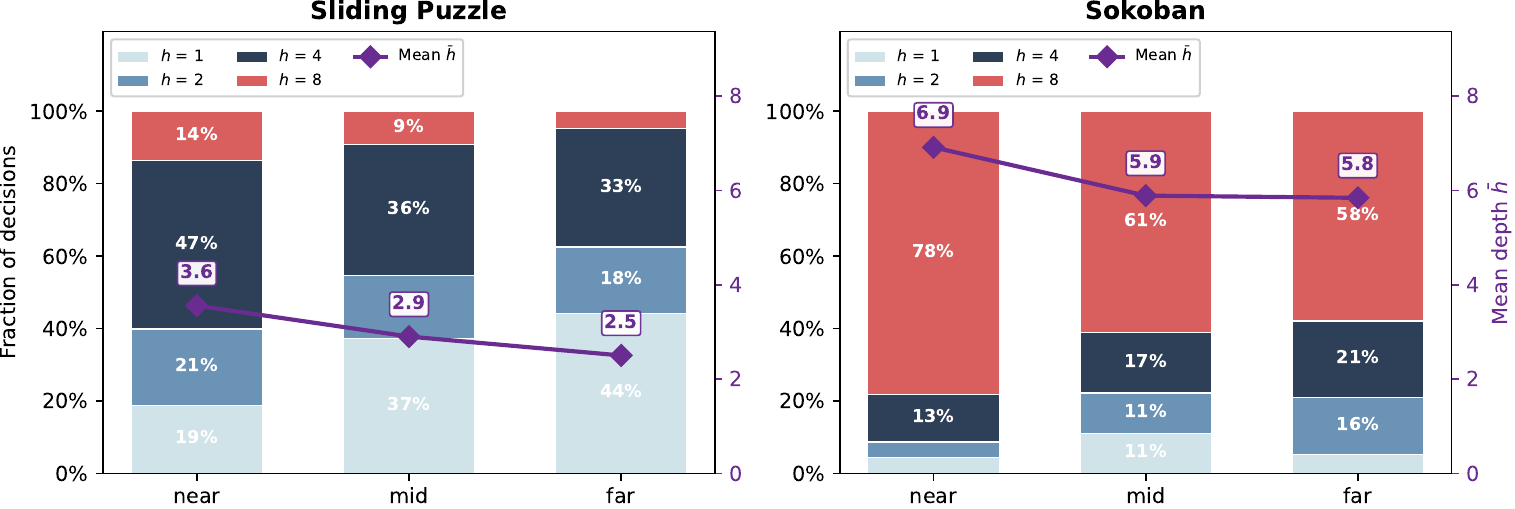}
    \caption{Chosen commitment depth versus remaining solver distance for the adaptive policy on Sliding and Sokoban, evaluated under a relaxed decision budget: the budget K at test time is larger than the budget used during training (budget-transfer setting; see the evaluation protocol in the paper). Stacked bars give the share of decisions at each chosen depth h within each distance bucket (near / mid / far, bucketed by solver distance at decision time); purple diamonds give the mean depth per bucket. Buckets with fewer than 15 decisions are hatched. The policy commits deep when close to the goal and shortens its commitments as the remaining distance grows.}
    \label{fig:h_frac}
\end{figure}

Figure~\ref{fig:h_frac} complements the matched-budget
analysis in Fig.~\ref{fig:h_frac_samek}.
With evaluation budgets exceeding training budgets,
both puzzles again exhibit greater mean selected
depth near the goal than farther away.
The distance-conditioned allocation pattern therefore
appears under both budget configurations.

\subsection{Depth distributions across settings}
\label{app:depth_distribution}
\begin{figure}[H]
    \centering
    \includegraphics[width=\linewidth]
    {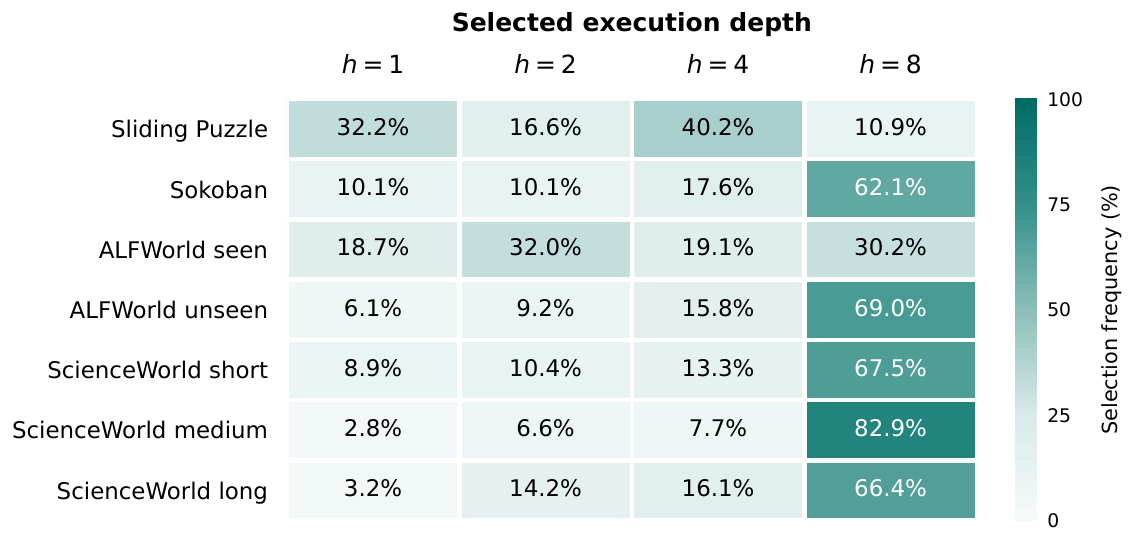}
    \caption{\textbf{Selected execution-depth distributions
    across seven settings.}
    Cells report selection frequencies (\%) for
    $h\in\{1,2,4,8\}$ under the main adaptive policy;
    darker shading indicates higher frequency.
    Values describe selected depths, not execution
    lengths truncated by episode termination.
    Percentages may sum to 99.9 or 100.1 due to rounding.}
    \label{fig:depth_distribution}
\end{figure}

Figure~\ref{fig:depth_distribution} summarizes
selected-depth frequencies under the main evaluation
configuration.
All four depths occur in every setting, although
their frequencies vary substantially.
ALFWorld seen distributes selections relatively
broadly, whereas ALFWorld unseen and all three
ScienceWorld groups favor $h=8$.
These aggregate distributions establish the use
of multiple depths; the puzzle analysis above
additionally relates depth choices to task progress.

\subsection{Failure Cases}
\label{app:failure_cases}

\paragraph{Open-loop simulation errors.} Because a chunk is executed without intermediate observations, the model must predict the effect of each of its own actions and plan the rest of the chunk on top of that prediction. The dominant failure in the visual puzzles is a mismatch between this internal simulation and the true dynamics, which compounds over the chunk. In Sokoban, such errors are often irreversible: a single miscounted push can drive a box against a wall or into a corner, after which no solution exists and the rest of the budget is spent in a dead state. Moves into walls are a direct symptom, since they reveal that the position the model assumed for its player differs from the true one. Figure~\ref{fig:teaser} (a) shows a representative case. The initial state is solvable in 10 moves, and the chunk ends with \texttt{U}, which pushes the upper box toward the top goal only if the player stands directly below it, i.e., after exactly two \texttt{L} moves. The model issues three. The third \texttt{L} pushes the lower box into a corner, making the puzzle unsolvable, and the final \texttt{U} walks into a wall. The model sees the outcome only after the entire chunk has executed, and all five remaining decisions are spent in an unsolvable state. In the Sliding Puzzle, failures are less abrupt but similar in kind: after a chunk the board frequently drifts away from the goal, and many failed episodes end farther from the solution than they started.

\begin{figure}[H]
    \centering
    \includegraphics[width=\linewidth]{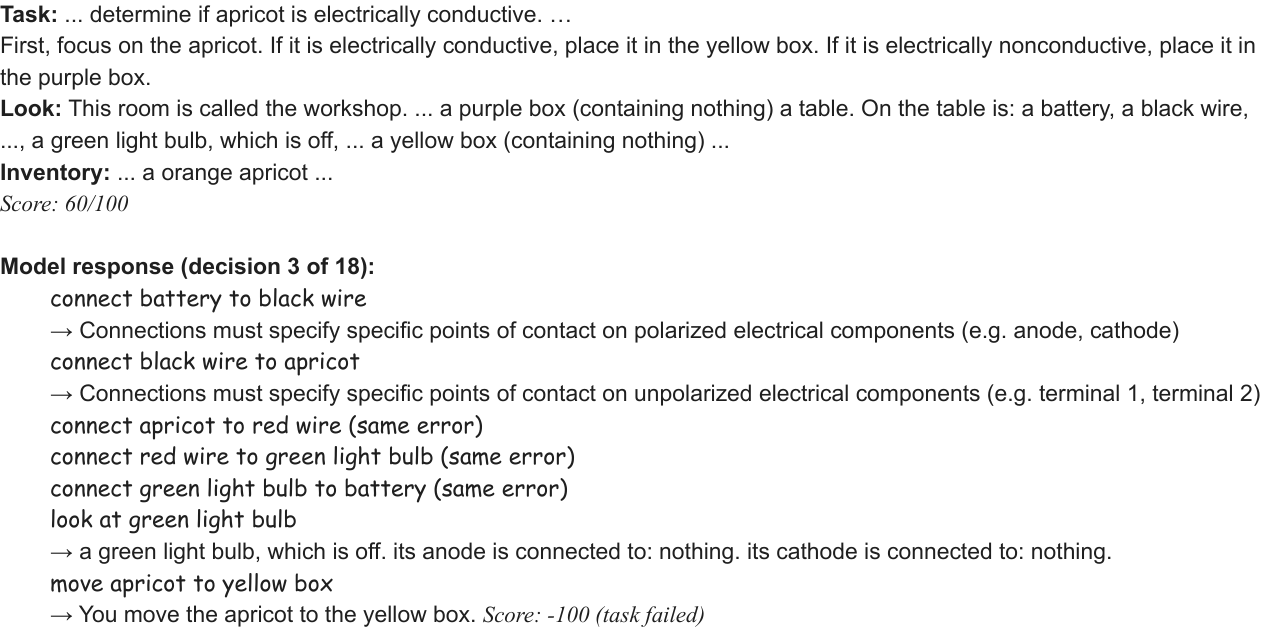}
    \caption{\textbf{Premature chunk in ScienceWorld}. The top block is an excerpt of the state the model received before its third decision. All actions below were proposed in a single chunk. Each simulator response shown was produced during execution but was not visible to the model when it chose the next action. The circuit is never built, yet the answer is executed in the same chunk. The chunk's eighth action (\texttt{look around}) was discarded when the episode ended.}
    \label{fig:failure_ScienceWorld}
\end{figure}

\paragraph{Premature chunk.} In ScienceWorld, a task is completed by a terminal action, such as focusing on an object or placing it in one of several answer boxes, and a wrong terminal action ends the episode with failure. Under open-loop execution, models frequently place the evidence-gathering step and the answer in the same chunk, so the answer is fixed before the evidence can be observed. Figure~\ref{fig:failure_ScienceWorld} shows such a case under Fixed-8. The model attempts to build a circuit to test whether an apricot conducts electricity, inspects the light bulb, and places the apricot in the box reserved for conductive substances, all within one chunk. None of the five \texttt{connect} commands succeeds, because each omits the terminal names the simulator requires. The bulb remains disconnected and off, yet the answer has already been scheduled. The episode ends with a score of $-100$ after only three of its 18 decisions, so the failure is caused by executing chunk too early, not by budget pressure. The same pattern appears when models focus on a substance and immediately on an answer box before any temperature has been read.

\paragraph{Other failures.} The remaining failures mostly exhaust the decision budget without reaching the goal. In ALFWorld, this typically reflects inefficient search, which open-loop execution makes worse. Models often script a fixed routine, such as \texttt{go to cabinet 4}, \texttt{open cabinet 4}, \texttt{take mug from cabinet 4}, before observing the receptacle. The \texttt{open} is then rejected because the receptacle is already open or has no door, and the \texttt{take} is rejected because the object is not there. Each rejected command still consumes a step. Weaker open models additionally struggle with the response format itself. Because a format-violating chunk is not executed and leaves the observation unchanged, such models tend to resubmit the identical response, consuming the budget without acting.

\section{Zero-Shot Depth Selection: Results and Protocol}
\label{app:zero_shot}

\subsection{Results}
\label{app:zero_shot_results}

We report success rates for prompted self-selected depth
and all four tested fixed depths, including conditions
that tie or outperform self-selected depth.
Comparisons are within each model and evaluation pool;
reasoning configurations and protocol differences are
documented in App.~\ref{app:zero_shot_protocol}.

\begin{table}[H]
\centering
\small
\setlength{\tabcolsep}{4.6pt} %
\renewcommand{\arraystretch}{1.12}
\caption{\textbf{Zero-shot execution-depth comparisons.}
Success rates (\%) for prompted self-selected depth
and fixed depths $h\in\{1,2,4,8\}$.
Self-selected depth is prompted rather than learned
through PACE training.
Reasoning configurations are model-specific;
these are not compute-matched cross-model comparisons.}
\label{tab:model_performance}
\begin{tabularx}{\columnwidth}{>{\raggedright\arraybackslash}p{4.15cm} l ccccc}
\toprule
\textbf{Model} & \textbf{Setting} & \textbf{Self-selected} & \textbf{Fixed-1} & \textbf{Fixed-2} & \textbf{Fixed-4} & \textbf{Fixed-8} \\
\midrule
Claude-Sonnet-5 & Sliding Puzzle & 3.1\%  & 0.0\% & 0.0\%  & 1.6\%  & 1.6\%  \\
\citep{anthropic_claude_2026}                & Sokoban        & 6.3\%  & 0.0\% & 3.1\%  & 7.8\%  & 10.9\% \\
\addlinespace[2.5pt]
Gemini-3.1-Pro-Preview & Sliding Puzzle & 0.0\%  & 0.0\% & 0.0\%  & 0.0\%  & 0.0\%  \\
\citep{geminiteam2025geminifamilyhighlycapable}                       & Sokoban        & 20.3\% & 0.0\% & 6.3\%  & 17.2\% & 18.8\% \\
\addlinespace[2.5pt]
GPT-5.6-Terra & Sliding Puzzle & 1.6\%  & 0.0\% & 0.0\%  & 4.7\%  & 6.3\%  \\
\citep{singh2026gpt5}              & Sokoban        & 57.8\% & 1.6\% & 17.2\% & 32.8\% & 43.8\% \\
\addlinespace[2.5pt]
Qwen2.5-VL-7B-Instruct & Sliding Puzzle & 0.0\%  & 0.0\% & 0.0\%  & 0.0\%  & 0.0\%  \\
\citep{bai2025qwen25vltechnicalreport}                       & Sokoban        & 0.0\%  & 0.0\% & 0.0\%  & 0.0\%  & 0.0\%  \\
\addlinespace[2.5pt]
Qwen3.8-27B & Sliding Puzzle & 4.7\%  & 0.0\% & 3.1\%  & 3.1\%  & 15.6\% \\
\citep{qwen38}            & Sokoban        & 18.8\% & 0.0\% & 10.9\% & 21.9\% & 10.9\% \\
\addlinespace[2.5pt]
Qwen3.8-Flash-Next & Sliding Puzzle & 21.9\% & 0.0\% & 4.7\%  & 34.4\% & 53.1\% \\
\citep{qwen3.8flashnext}                   & Sokoban        & 28.1\% & 1.6\% & 17.2\% & 18.8\% & 34.4\% \\
\addlinespace[2.5pt]
GLM-5.3-Flash & Sliding Puzzle & 26.6\% & 0.0\% & 10.9\% & 28.1\% & 17.2\% \\
\citep{glm5team2026glm5vibecodingagentic}              & Sokoban        & 17.2\% & 1.6\% & 15.6\% & 20.3\% & 26.6\% \\
\addlinespace[2.5pt]
Gemma-4-31B & Sliding Puzzle & 4.7\%  & 0.0\% & 0.0\%  & 0.0\%  & 0.0\%  \\
\citep{gemmateam2026gemma4}            & Sokoban        & 34.4\% & 1.6\% & 12.5\% & 26.6\% & 31.2\% \\
\addlinespace[2.5pt]
DeepSeek-V4-Flash-Vision-Exp & Sliding Puzzle & 0.0\%  & 0.0\% & 0.0\%  & 0.0\%  & 0.0\%  \\
\citep{deepseek2026v4flashvisionexp}                             & Sokoban        & 10.9\% & 0.0\% & 4.7\%  & 6.2\%  & 10.9\% \\
\bottomrule
\end{tabularx}
\end{table}

\begin{table}[H]
\centering
\small
\setlength{\tabcolsep}{6.3pt} %
\renewcommand{\arraystretch}{1.12}
\caption{\textbf{ALFWorld Zero-shot execution-depth comparisons.}
Success rates (\%) for prompted self-selected depth
and fixed depths $h\in\{1,2,4,8\}$.
Self-selected depth is prompted rather than learned
through PACE training.
Reasoning configurations are model-specific;
these are not compute-matched cross-model comparisons.}
\label{tab:zeroshot_alfworld}
\begin{tabularx}{\columnwidth}{>{\raggedright\arraybackslash}p{4.15cm} l ccccc}
\toprule
\textbf{Model} & \textbf{Setting} & \textbf{Self-selected} & \textbf{Fixed-1} & \textbf{Fixed-2} & \textbf{Fixed-4} & \textbf{Fixed-8} \\
\midrule
Claude-Sonnet-5 & All & 94.8\% & 81.5\% & 92.7\% & 92.3\% & 92.7\% \\
\citep{anthropic_claude_2026} & seen & 94.8\% & 85.3\% & 94.8\% & 91.4\% & 89.7\% \\
 & unseen & 94.9\% & 77.8\% & 90.6\% & 93.2\% & 95.7\% \\
\addlinespace[2.5pt]
Gemini-3.1-Pro-Preview & All & 97.9\% & 78.1\% & 91.4\% & 97.4\% & 98.7\% \\
\citep{geminiteam2025geminifamilyhighlycapable} & seen & 96.6\% & 81.0\% & 90.5\% & 94.8\% & 97.4\% \\
 & unseen & 99.1\% & 75.2\% & 92.3\% & 100.0\% & 100.0\% \\
\addlinespace[2.5pt]
GPT-5.6-Terra & All & 91.4\% & 45.5\% & 64.8\% & 85.4\% & 95.7\% \\
\citep{singh2026gpt5} & seen & 89.7\% & 44.0\% & 58.6\% & 81.9\% & 95.7\% \\
 & unseen & 93.2\% & 47.0\% & 70.9\% & 88.9\% & 95.7\% \\
\addlinespace[2.5pt]
Qwen2.5-VL-7B-Instruct & All & 4.3\% & 12.9\% & 11.2\% & 2.6\% & 4.3\% \\
\citep{bai2025qwen25vltechnicalreport} & seen & 5.2\% & 13.8\% & 14.7\% & 4.3\% & 6.9\% \\
 & unseen & 3.4\% & 12.0\% & 7.7\% & 0.9\% & 1.7\% \\
\addlinespace[2.5pt]
Qwen3.8-27B & All & 91.8\% & 48.1\% & 79.8\% & 89.3\% & 96.1\% \\
\citep{qwen38} & seen & 93.1\% & 46.6\% & 77.6\% & 85.3\% & 95.7\% \\
 & unseen & 90.6\% & 49.6\% & 82.1\% & 93.2\% & 96.6\% \\
\addlinespace[2.5pt]
Qwen3.8-Flash-Next & All & 88.8\% & 63.9\% & 79.8\% & 91.4\% & 98.3\% \\
\citep{qwen3.8flashnext} & seen & 92.2\% & 62.9\% & 83.6\% & 93.1\% & 97.4\% \\
 & unseen & 85.5\% & 65.0\% & 76.1\% & 89.7\% & 99.1\% \\
\addlinespace[2.5pt]
GLM-5.3-Flash & All & 90.6\% & 56.2\% & 78.5\% & 87.6\% & 75.1\% \\
\citep{glm5team2026glm5vibecodingagentic} & seen & 88.8\% & 56.0\% & 75.9\% & 86.2\% & 78.4\% \\
 & unseen & 92.3\% & 56.4\% & 81.2\% & 88.9\% & 71.8\% \\
\addlinespace[2.5pt]
Gemma-4-31B & All & 73.0\% & 41.2\% & 56.7\% & 62.7\% & 58.4\% \\
\citep{gemmateam2026gemma4} & seen & 72.4\% & 41.4\% & 51.7\% & 58.6\% & 53.4\% \\
 & unseen & 73.5\% & 41.0\% & 61.5\% & 66.7\% & 63.2\% \\
\addlinespace[2.5pt]
DeepSeek-V4-Flash-Vision-Exp & All & 77.7\% & 35.6\% & 48.1\% & 76.0\% & 91.8\% \\
\citep{deepseek2026v4flashvisionexp} & seen & 72.4\% & 31.0\% & 49.1\% & 73.3\% & 88.8\% \\
 & unseen & 82.9\% & 40.2\% & 47.0\% & 78.6\% & 94.9\% \\
\bottomrule
\end{tabularx}
\end{table}

\begin{table}[H]
\centering
\small
\setlength{\tabcolsep}{6.2pt} %
\renewcommand{\arraystretch}{1.12}
\caption{\textbf{ScienceWorld Zero-shot execution-depth comparisons.}
Success rates (\%) for prompted self-selected depth
and fixed depths $h\in\{1,2,4,8\}$.
Self-selected depth is prompted rather than learned
through PACE training.
Reasoning configurations are model-specific;
these are not compute-matched cross-model comparisons.}
\label{tab:zeroshot_ScienceWorld}
\begin{tabularx}{\columnwidth}{>{\raggedright\arraybackslash}p{4.15cm} l ccccc}
\toprule
\textbf{Model} & \textbf{Setting} & \textbf{Self-selected} & \textbf{Fixed-1} & \textbf{Fixed-2} & \textbf{Fixed-4} & \textbf{Fixed-8} \\
\midrule
Claude-Sonnet-5 & All & 54.1\% & 36.5\% & 49.2\% & 64.6\% & 61.3\% \\
\citep{anthropic_claude_2026} & short & 31.3\% & 10.4\% & 25.4\% & 58.2\% & 56.7\% \\
 & medium & 51.9\% & 31.5\% & 42.6\% & 53.7\% & 63.0\% \\
 & long & 81.7\% & 70.0\% & 81.7\% & 81.7\% & 65.0\% \\
\addlinespace[2.5pt]
Gemini-3.1-Pro-Preview & All & 78.5\% & 38.7\% & 63.0\% & 77.3\% & 81.2\% \\
\citep{geminiteam2025geminifamilyhighlycapable} & short & 73.1\% & 14.9\% & 40.3\% & 68.7\% & 70.1\% \\
 & medium & 81.5\% & 38.9\% & 72.2\% & 79.6\% & 87.0\% \\
 & long & 81.7\% & 65.0\% & 80.0\% & 85.0\% & 88.3\% \\
\addlinespace[2.5pt]
GPT-5.6-Terra & All & 68.0\% & 40.9\% & 56.4\% & 72.9\% & 64.1\% \\
\citep{singh2026gpt5} & short & 46.3\% & 20.9\% & 31.3\% & 58.2\% & 52.2\% \\
 & medium & 72.2\% & 33.3\% & 51.9\% & 74.1\% & 66.7\% \\
 & long & 88.3\% & 70.0\% & 88.3\% & 88.3\% & 75.0\% \\
\addlinespace[2.5pt]
Qwen2.5-VL-7B-Instruct & All & 1.1\% & 0.6\% & 0.6\% & 0.0\% & 0.0\% \\
\citep{bai2025qwen25vltechnicalreport} & short & 0.0\% & 0.0\% & 0.0\% & 0.0\% & 0.0\% \\
 & medium & 1.9\% & 0.0\% & 1.9\% & 0.0\% & 0.0\% \\
 & long & 1.7\% & 1.7\% & 0.0\% & 0.0\% & 0.0\% \\
\addlinespace[2.5pt]
Qwen3.8-27B & All & 71.8\% & 40.9\% & 60.8\% & 66.9\% & 65.2\% \\
\citep{qwen38} & short & 59.7\% & 10.4\% & 37.3\% & 55.2\% & 49.3\% \\
 & medium & 75.9\% & 48.1\% & 70.4\% & 63.0\% & 64.8\% \\
 & long & 81.7\% & 68.3\% & 78.3\% & 83.3\% & 83.3\% \\
\addlinespace[2.5pt]
Qwen3.8-Flash-Next & All & 65.7\% & 36.5\% & 60.2\% & 68.5\% & 64.6\% \\
\citep{qwen3.8flashnext} & short & 49.3\% & 14.9\% & 41.8\% & 53.7\% & 44.8\% \\
 & medium & 70.4\% & 48.1\% & 63.0\% & 70.4\% & 64.8\% \\
 & long & 80.0\% & 50.0\% & 78.3\% & 83.3\% & 86.7\% \\
\addlinespace[2.5pt]
GLM-5.3-Flash & All & 40.3\% & 22.1\% & 43.1\% & 53.0\% & 49.2\% \\
\citep{glm5team2026glm5vibecodingagentic} & short & 16.4\% & 7.5\% & 29.9\% & 32.8\% & 28.4\% \\
 & medium & 42.6\% & 22.2\% & 42.6\% & 51.9\% & 48.1\% \\
 & long & 65.0\% & 38.3\% & 58.3\% & 76.7\% & 73.3\% \\
\addlinespace[2.5pt]
Gemma-4-31B & All & 49.7\% & 24.3\% & 33.7\% & 58.6\% & 47.5\% \\
\citep{gemmateam2026gemma4} & short & 35.8\% & 7.5\% & 19.4\% & 52.2\% & 31.3\% \\
 & medium & 51.9\% & 33.3\% & 31.5\% & 55.6\% & 44.4\% \\
 & long & 63.3\% & 35.0\% & 51.7\% & 68.3\% & 68.3\% \\
\addlinespace[2.5pt]
DeepSeek-V4-Flash-Vision-Exp & All & 39.8\% & 27.1\% & 39.2\% & 45.3\% & 49.7\% \\
\citep{deepseek2026v4flashvisionexp} & short & 16.4\% & 10.4\% & 20.9\% & 28.4\% & 38.8\% \\
 & medium & 46.3\% & 31.5\% & 38.9\% & 48.1\% & 50.0\% \\
 & long & 60.0\% & 41.7\% & 60.0\% & 61.7\% & 61.7\% \\
\bottomrule
\end{tabularx}
\end{table}

\subsection{Evaluation Protocol}
\label{app:zero_shot_protocol}

\paragraph{Purpose and scope.}
This evaluation asks whether prompting a pretrained
model to select its execution depth improves performance
over prescribing a constant depth.
No task-specific parameter updates are performed.
We distinguish prompted self-selected depth from
PACE's trained depth head.
These experiments provide supplementary evidence about
execution timing, rather than a matched ranking against
PACE: the systems differ in training, action
representation, context handling, and reasoning configuration.
Moreover, changing the requested chunk length can
change the generated actions themselves.
The prompted conditions therefore do not isolate timing
in the same way as PACE's same-checkpoint depth overrides.

\paragraph{Depth conditions and decision budgets.}
Each evaluation entry is tested under five conditions.
In the self-selected condition, the model chooses
$h\in\mathcal{H}=\{1,2,4,8\}$ at each decision.
In Fixed-$h$, it is instructed to return exactly $h$ actions.
The best fixed depth is selected in hindsight from
these four values within the reported evaluation pool,
rather than the eight depths tested in PACE's
main comparisons.

Decision budgets are $K = 15$ for Sliding Puzzle, $K = 6$ for Sokoban, $K = 9$ for
ALFWorld, and $K = 6/9/18$ for ScienceWorld short/medium/long, matching the main
experiments. Every returned response consumes one decision, including responses
rejected for format violations.

\paragraph{Interaction and observations.}
Each episode is a multi-turn interaction.
At a decision boundary, the model receives the current
observation and returns an action chunk.
The chunk executes open-loop, without intermediate
feedback; new information is supplied at the next
decision boundary.

The puzzle interface uses rendered board observations.
For ALFWorld and ScienceWorld, observations are structured
text summaries accompanied by the preceding chunk's
actions and environment responses.
These responses become available only after execution
and cannot alter actions already selected within
the chunk.
Conversation history is retained across decisions.
Provider-supported reasoning content or continuation
metadata is preserved in its supported format.

Within each environment, prompt templates share the
task description and observation format, while depth
instructions and response schemas depend on the condition.
Prompts specify action semantics and output requirements
without providing solved demonstrations.
Examples appear in App.~\ref{app:zero_shot_prompts}.

\paragraph{Structured responses and validation.}
The requested JSON schemas are
\texttt{\{"h": int, "actions": [str]\}} for
self-selected depth,
\texttt{\{"action": str\}} for Fixed-1, and
\texttt{\{"actions": [str]\}} for other fixed depths.
Puzzle actions are drawn from \texttt{\{U,D,L,R\}};
text-domain actions are generated commands whose
validity is determined by the environment.

A response is rejected if its action list is missing
or ill-typed, if its declared depth disagrees with
the list length, or if a fixed-depth response contains
the wrong number of actions.
Rejected responses execute no actions but still
consume a decision.
The model receives no additional corrective error message.

For a self-consistent positive depth outside
$\mathcal{H}$, the executor retains the prefix of length
$\max\{h'\in\mathcal{H}:h'\leq h\}$.
Clipping and format violations are logged separately.
Observed performance therefore reflects both task
execution and compliance with the requested interface.

\paragraph{Execution and termination.}
Actions in an accepted chunk are passed to the
environment in order.
Nonterminal invalid commands consume environment steps
without triggering replanning.
Execution stops when the environment terminates,
a configured environment-step limit is reached,
or the decision budget is exhausted.
Any unexecuted chunk suffix is discarded.

\paragraph{Reasoning and serving configurations.}
Closed models use their providers' low reasoning-effort
settings to limit evaluation cost.
Open-weight models use the highest reasoning setting
available in the evaluated serving configuration.
These settings are not calibrated to equal token,
latency, or computational budgets.
The results therefore do not establish a compute-matched
ranking across models or between closed and open-weight
systems.

Within each model, the reasoning and decoding
configuration is held fixed across depth conditions.
Closed models use provider-native structured-output
interfaces; open-weight models use the serving
framework's guided decoding.
These mechanisms share the requested schemas but
need not induce identical decoding behavior.

\paragraph{Vocabulary coverage.}
Episodes flagged because their expert demonstrations
contain commands outside PACE's vocabulary remain
in the primary zero-shot evaluation pools and are
scored by actual rollout outcomes.
For text-generating models, this flag describes
reference-path coverage relative to PACE's vocabulary;
it does not restrict their own command support.

\paragraph{Aggregation and interpretation.}
Where results are pooled across splits or horizon
groups, success is computed from pooled success
counts and episode counts.
The best fixed depth for a pooled result is selected
after pooling each fixed-depth condition.
This differs from selecting a separate best fixed
depth within every split or horizon group.

All evaluated models and depth conditions are reported,
including ties, near-zero results, and cases where
self-selected depth underperforms fixed depth.
High or low success under this protocol does not,
by itself, identify the contribution of pretraining,
reasoning effort, or interface constraints.

\paragraph{Fault handling and logging.}
Rate-limit rejections are retried without consuming
a decision.
Infrastructure failures that prevent obtaining a
response trigger recollection of the episode;
parseable responses that violate the action schema
are charged under the validation rules above.

Logs retain model identifiers, inference configurations,
prompts, raw responses, requested and executed actions,
validation outcomes, environment feedback, token usage,
and latency.
These records distinguish failures of the evaluated
policy interface from serving-infrastructure failures.

\subsection{Prompt Templates}
\label{app:zero_shot_prompts}

\begin{figure}[H]
    \centering
    \includegraphics[width=1\linewidth]{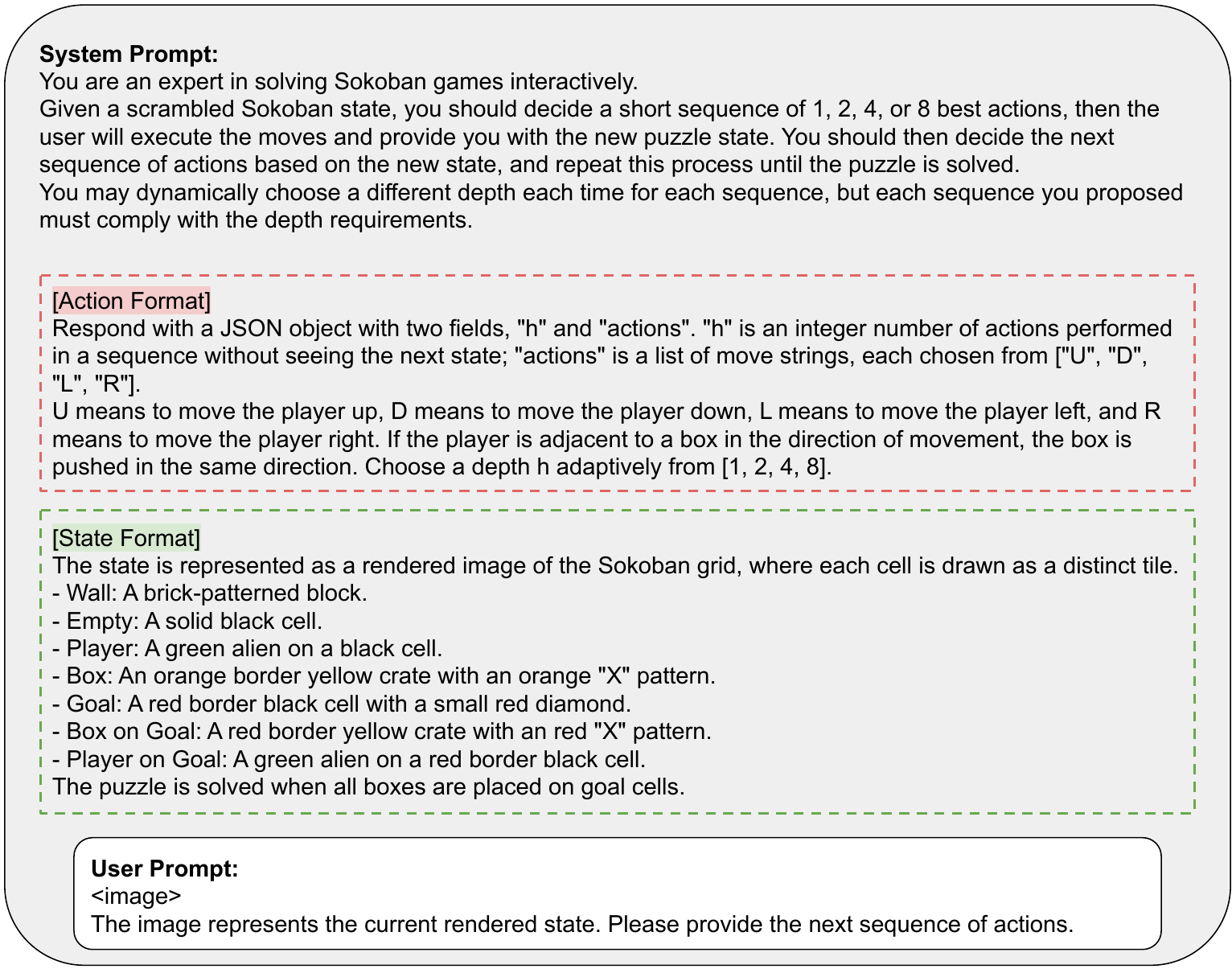}
    \caption{\textbf{Example system prompt for zero-shot vision-language model baseline on Sokoban (image state).} The puzzle name and its state and action format descriptions are task-dependent and are different for different tasks.}
    \label{fig:prompt_example1}
\end{figure}

The prompt template examples in Fig.~\ref{fig:prompt_example1} specify the task, observation format, action semantics, and execution-depth condition.

\section{Limitations}

Our experiments use Qwen2.5-VL-7B-Instruct~\citep{bai2025qwen25vltechnicalreport};
evaluation with other backbones remains future work.
We also focus on discrete-action environments.
The principle of learned execution depth extends conceptually
to continuous control, but applying PACE to broader
robotic learning settings requires adapting the action
decoder, backbone integration, and training recipe.
We leave these adaptations and their empirical validation
to future work.

\end{document}